\documentclass[preprint,12pt,authoryear]{elsarticle}

\usepackage{amssymb}
\usepackage{amsmath}

\usepackage{graphicx}
\usepackage{hyperref}
\usepackage{enumitem}
\usepackage{algorithm}
\usepackage{algpseudocode}
\usepackage{booktabs}
\usepackage{caption}
\usepackage{siunitx}
\usepackage{orcidlink}
\usepackage[margin=1in]{geometry}
\usepackage{siunitx}

\journal{}

\begin{document}

\begin{frontmatter}

\title{Foam Segmentation in Wastewater Treatment Plants: A Federated Learning Approach with Segment Anything Model 2}

\author[cor1]{Mehmet Batuhan Duman\orcidlink{0009-0007-1249-9178}}
\ead{batuhanduman@uma.es}

\author{Alejandro Carnero}
\ead{alexcarnero97@uma.es}

\author{Cristian Mart\'in}
\ead{cristian@uma.es}

\author{Daniel Garrido}
\ead{dgm@uma.es}

\author{Manuel~D\'iaz}
\ead{mdiaz@uma.es}

\cortext[cor1]{Corresponding author}

\affiliation{organization={ITIS Software, University of Malaga},city={Malaga},country={Spain}}

\begin{abstract}
Foam formation in Wastewater Treatment Plants (WTPs) is a major challenge that can reduce treatment efficiency and increase costs. The ability to automatically examine changes in real-time with respect to the percentage of foam can be of great benefit to the plant. However, large amounts of labeled data are required to train standard Machine Learning (ML) models. The development of these systems is slow due to the scarcity and heterogeneity of labeled data. Additionally, the development is often hindered by the fact that different WTPs do not share their data due to privacy concerns. This paper proposes a new framework to address these challenges by combining Federated Learning (FL) with the state-of-the-art base model for image segmentation, Segment Anything Model 2 (SAM2). The FL paradigm enables collaborative model training across multiple WTPs without centralizing sensitive operational data, thereby ensuring privacy. The framework accelerates training convergence and improves segmentation performance even with limited local datasets by leveraging SAM2's strong pre-trained weights for initialization. The methodology involves fine-tuning SAM2 on distributed clients (edge nodes) using the Flower framework, where a central Fog server orchestrates the process by aggregating model weights without accessing private data. The model was trained and validated using various data collections, including real-world images captured at a WTPs in Granada, Spain, a synthetically generated foam dataset, and images from publicly available datasets to improve generalization. This research offers a practical, scalable, and privacy-aware solution for automatic foam tracking in WTPs. The findings highlight the significant potential of integrating large-scale foundational models into FL systems to solve real-world industrial challenges characterized by distributed and sensitive data.
\end{abstract}




\end{frontmatter}

\section{Introduction}
\label{sec:introduction}

Foaming in Wastewater Treatment Plants (WTPs) is a significant operational challenge that reduces treatment efficiency and increases operating costs. The presence of foam in aeration tanks and clarifiers leads to reduced oxygen transfer, decreased biomass concentration, odor problems, and increased maintenance expenses \cite{collivignarelli2020foams, ruzicka2009cause}. Therefore, accurate foam detection and quantification are essential for implementing timely control measures and optimizing WTPs performance.

Foam generation in WTPs arises from a complex interplay of factors, including surfactants and detergents present in influent wastewater, excessive growth of filamentous bacteria, imbalances within the biological treatment process, and the presence of proteins and lipids \cite{wang2013enhancing}. Filamentous bacteria such as \textit{Microthrix parvicella} and \textit{Gordonia} species produce biosurfactants that stabilize foams \cite{petrovski2011examination}. If not addressed, uncontrolled foam significantly compromises treatment efficacy, raises safety concerns, and risks the discharge of partially treated effluent into the environment.

Traditionally, foam monitoring has relied on visual inspection or basic sensor systems, offering limited precision and objectivity. Advanced image analysis techniques promise more accurate, automated, and real-time foam segmentation but remain challenged by data privacy, heterogeneous foam characteristics between facilities, scarce labeled data, and the requirement for low-latency processing \cite{Zhang2020}. Recent studies in diverse domains underscore similar hurdles: distributed data with privacy constraints in medical imaging \cite{subedi2023client, peketi2023flwgan}, water-level prediction \cite{thonglek2024hierarchical}, and agricultural irrigation \cite{ahmadi2024federated} all highlighting the need for collaborative, privacy-aware methods to manage non-identical data distributions.

Deep learning approaches, notably Convolutional Neural Networks (CNNs), have proven effective for image segmentation in fields such as medical diagnosis \cite{ronneberger2015u} and industrial defect detection \cite{li2022detection, gu2022research}. However, directly applying deep learning methods to WTPs foam segmentation is complicated by sensitive operational data and the wide variability in plant designs \cite{batinovic2021cocultivation}. Federated Learning (FL) is emerging as a viable solution: a distributed machine learning paradigm that enables training models across geographically dispersed systems. Models are trained locally on decentralized devices (or servers) without sharing their private data, thus maintaining privacy \cite{mcmahan2017communication}. FL has already shown promise in multiple use cases, including pothole detection \cite{singh2024enhanced}, sustainable transportation \cite{zhou2024adaptive}, and remote sensing \cite{zhu2023privacy}, but these efforts often face slow convergence or reduced performance under non-Independent and Identically Distributed (non-IID) data conditions.

This paper primarily focuses on proposing a novel framework for foam detection that leverages the advantages of FL combined with initialization from a large-scale foundation model. Specifically, we harness the Segment Anything Model 2 (SAM2) \cite{ravi2024sam}, whose broad segmentation capabilities have been successfully adapted to specialized domains \cite{mehrnia2024assessing}. Leveraging SAM2’s pre-trained weights accelerates training convergence and increases performance, particularly where labeled data are scarce. In this context, we further explore SAM2, a hierarchical variant of SAM designed for multi-scale feature extraction and point-based prompting. Together, these models provide a robust foundation for foam segmentation in real-world WTPs conditions.

This work contributes significantly to the wastewater treatment domain by delivering a privacy-preserving, high-performance foam segmentation strategy adaptable to diverse plant conditions. Beyond WTPs, the insights gained may extend to other resource management and industrial applications where data sharing is restricted and produced in real-time, where distributed learning is crucial. Similar insights can be drawn from knowledge distillation approaches, which reduce communication overhead in federated scenarios \cite{sun2024fkd}, but our focus is on leveraging SAM/SAM2 initialization for more resilient foam segmentation. Our method features three main components:

\begin{itemize} 
    \item A FL framework tailored for WTPs foam segmentation.
    \item A novel initialization strategy using SAM2, adapted to the unique characteristics of foam images.
    \item A systematic evaluation under both IID and non-IID data splits, reflecting realistic WTPs heterogeneity.
\end{itemize}

Experiments on real WTPs data demonstrate that FL models initialized with SAM2 converge in a significantly smaller number of communication rounds compared to models with random initialization.  Our SAM2-based approach achieves a Dice score of 84.38\% using FL with two clients. The hierarchical architecture of SAM2 further contributes to the refinement of segmentation boundaries, facilitating accurate foam detection under difficult conditions.

While classical segmentation models such as U-Net and DeepLabV3+ can achieve higher scores on certain metrics in centralized training, SAM2 is shown to provide highly competitive and robust performance within the practical constraints of a federated, privacy-preserving system. The reasons for these performance characteristics are explored in our prompt-sensitivity analysis, which shows that despite differences in evaluation scores, SAM2's visual outputs are comparable to other leading models.

This paper details the fine-tuning process of SAM2 for foam segmentation, the methodology for creating the specialized training dataset, the implementation of the federated learning framework across different clients, and the final deployment and evaluation in real-world WTPs.

The present study begins with a contextualization of the research within the existing literature on foam segmentation and distributed learning. Next, we delineate our primary contribution: a novel framework integrating disparate datasets and a federated learning methodology based on the SAM2 model. We also provide a detailed exposition of its practical implementation. The paper concludes with a comprehensive performance evaluation, a discussion of the project's challenges, and a summary of our findings and future research directions.

\section{Related Work}
\label{sec:relatedwork}

The formation and detection of foam in WTPs have been the subject of research for several decades. Researchers have investigated foam formation from various angles, including understanding the underlying causes and developing advanced detection and monitoring techniques. Across domains such as manufacturing \cite{li2022detection}, water conservation \cite{ahmadi2024federated}, and river segmentation \cite{yu2020segmentation}, the challenge often lies in handling heterogeneous data and environmental factors that reflect the complexity of the foam characteristics in WTPs.

Investigations focused on identifying key factors contributing to foam formation. A comprehensive study by \cite{wang2013enhancing} highlighted the complex interplay between biological and physicochemical processes in foam formation. They identified surfactant concentration, filamentous bacterial growth, and process imbalances as critical factors. More detailed investigation was conducted by \cite{petrovski2011examination} into the specific roles of various microorganisms, with a particular focus on Microthrix parvicella and several Mycolata species. Their research provided valuable insights into the metabolic pathways and surface properties of these organisms, furthering our understanding of the biological aspects of foam formation in WTPs.

The physicochemical properties of WTPs foam have been studied in \cite{Karakashev2012}. Their work on foam stability and dewatering properties provided a foundation for understanding the behavior of WTPs foams under different operating conditions. These findings have proven critical for the development of effective foam detection and control strategies. Similarly, it was shown by \cite{gu2022research} that even in other porous materials, such as carbon foam, segmentation algorithms must account for material-specific properties and noise, suggesting that dedicated adaptations may be required for foam monitoring in WTPs.

The application of image analysis to foam segmentation represented a significant leap forward in monitoring capabilities. In \cite{ruzicka2009cause}, an image processing algorithm specifically designed for foam detection in the context of wastewater treatment was developed. Their approach, which combined thresholding techniques with morphological operations, showed promise in accurately classifying foam regions in digital images. Recent advancements in other industrial settings, such as nickel foam defect detection \cite{li2022detection}, reinforce the importance of lightweight yet effective segmentation models for real-time quality control and monitoring, a requirement that also applies to foam detection in WTPs.

The advent of deep learning algorithms has precipitated a paradigm shift in the domain of image segmentation, encompassing diverse disciplines such as foam tracking of WTPs.  In \cite{ronneberger2015u}, the U-Net architecture, which has since become a cornerstone of many segmentation tasks, is introduced. U-Net's ability to capture local and global features makes it ideal for complex segmentation.\cite{long2015fully} proposed a fully convolutional network for semantic segmentation and demonstrated state-of-the-art performance on various benchmarks, laying the foundation for end-to-end trainable segmentation models. In wastewater treatment, \cite{carballo2024foam} successfully employed deep models such as DeepLabv3+ \cite{chen2018encoder} and One Shot Texture Segmentation (OSTS) to achieve a Dice score of 86\%, demonstrating the feasibility of advanced computer vision techniques in this application. However, challenges remain in scaling these methods to diverse facilities while preserving operational data privacy.

As the field evolves, there is a growing recognition of the need for privacy-preserving approaches to WTPs data analysis. FL has emerged as a powerful paradigm for distributed ML that addresses many of the privacy and data-sharing concerns inherent in centralized data collection \cite{mcmahan2017communication}. FL has been applied in numerous contexts, including medical imaging \cite{subedi2023client, peketi2023flwgan}, transportation \cite{zhou2024adaptive}, and water resource management \cite{thonglek2024hierarchical}, consistently highlighting its potential to aggregate knowledge across disparate data silos. Such decentralized strategies are closely aligned with the confidentiality requirements of wastewater treatment, where the centralization of raw data can be difficult or impossible.

The use of FL in the context of WTPs foam monitoring has been explored by \cite{carballo2024foam}. Although their specific case did not show significant improvements over centralized training due to the similarity of the client datasets, it highlighted the promise of privacy-preserving, distributed models when applied to heterogeneous WTPs. Subsequent work in federated scenarios has demonstrated that methods such as knowledge distillation \cite{sun2024fkd} or domain-specific model partitioning \cite{subedi2023client} can help mitigate non-IID data challenges, reduce communication overhead, and improve convergence speed.

The concept of foundation models, large-scale pre-trained networks adaptable to various tasks, has also shown promise in image segmentation. \cite{kirillov2023segment} introduced the Segment Anything Model (SAM), a versatile, large-scale model for segmentation. SAM2’s generalization ability makes it an appealing candidate for challenging domains like wastewater foam segmentation. Related research, such as \cite{mehrnia2024assessing}, further indicates that specialized variants of such models (e.g., MedSAM) can significantly accelerate and improve segmentation in low-data or privacy-sensitive settings. Transfer learning approaches \cite{raghu2019transfusion} have similarly demonstrated that robust pre-trained representations can reduce dependence on large annotated datasets.

Despite these developments, significant challenges persist in analyzing operational data from Advanced Aeration Tanks (AATs) and other WTPs processes. In \cite{driver2024encrypted}, issues of data quality and reliability in AATs monitoring were highlighted. Privacy considerations were further underscored by \cite{newhart2019data}, who examined the legal and ethical implications of sharing WTPs operational data. Also, the process conditions are always changing, and there are not many large labeled datasets. This shows that a complete approach that includes foam detection, deep learning, and privacy protections is needed.

In conclusion, while significant progress has been made in foam detection, image segmentation, and distributed learning, an integrated approach tailored to the specific demands of WTPs foam segmentation is still lacking. The work in \cite{carballo2024foam} marks an important step by merging advanced computer vision with FL-based privacy preservation. Nonetheless, there is ample room for improvements in model generalization, adaptability to diverse operational settings, and real-time processing capabilities. Future research should prioritize the development of robust, efficient frameworks capable of handling non-IID data distributions while maintaining high accuracy and strong privacy protections. The integration of these techniques into automated control systems may lead to more responsive and cost-effective WTPs, ultimately enhancing water treatment processes and environmental outcomes.

\section{Datasets and Pre-processing}
\label{sec:datasets}

Effective foam segmentation requires both diverse and well-managed images to capture the many ways foam manifests itself, so that sensitive plant data does not leave local facilities. To meet this dual demand, we identified three complementary methods:

\begin{table}[h!]
\centering
\caption{Summary of Datasets Used for Training and Evaluation.}
\label{tab:dataset_summary}
\begin{tabular}{@{}lcr@{}}
\toprule
\textbf{Dataset} & \textbf{Description} & \textbf{Images} \\
\midrule
ADE20K (Water) & Open-source water segmentations for zero-shot transfer & 1,888 \\
Granada WTP      & Real-world images from a WTP in Granada, Spain & 1,792 \\
Synthetic Foam   & Synthetic images of rare foam    & 300   \\
\midrule
\textbf{Total}   &  & \textbf{3,980} \\
\bottomrule
\end{tabular}
\end{table}

\begin{itemize}
    \item An open-source water dataset that enables zero-shot transfer and provides extensive visual prior information.
    \item A diffusion-based synthetic dataset that intentionally oversamples rare or hard-to-capture foam patterns.
    \item Time-stamped photographs collected under a data sharing agreement from a full-scale WTPs in Granada (Spain).
\end{itemize}

Before training, all images from these sources are processed through a uniform pipeline. This includes resizing and normalizing images for SAM2, generating prompts (points or boxes) from the masks, and applying augmentations to simulate plant conditions. The following subsections detail these steps and the final dataset splits.

\subsection{Data Preprocessing and Augmentation}

In order to enhance model robustness and address the limitations associated with scarce and heterogeneous datasets in WTPs, an extensive data preprocessing and augmentation protocol is implemented:

\begin{enumerate}[noitemsep]
    \item \textbf{Resizing:} All images are resized to 1024×1024 pixels to meet SAM2’s input requirements. This standardization aligns with SAM2’s initial training and model architecture, and ensures consistent input sizes across all clients.
    \item \textbf{Mask Creation:} Since the manual creation of segmented foam masks was both time-consuming and prone to error, masks were generated using a hybrid approach that combined classical image-processing algorithms with SAM2’s zero-shot predictions. This approach yielded stable, high-quality masks without requiring extensive manual annotation.
    \item \textbf{Image Augmentation:} To further diversify the training data and mitigate overfitting, a suite of geometric and photometric transformations was applied using the Albumentations library\cite{info11020125}. Specifically, each image mask pair was randomly subjected to:
      \begin{itemize}[noitemsep]
        \item Horizontal and vertical flips (\(p=0.5\) each).
        \item Random brightness and contrast adjustments (\(p=0.2\)).
        \item Random shifts (±5\%), scales (±10\%), and rotations (±15°) (\(p=0.5\)).
      \end{itemize}
      After augmentation, images and masks are converted to tensors with \texttt{ToTensorV2()}, normalized to \([0,1]\), and permuted to channel‐first format.
\end{enumerate}

\subsection{Data Labeling and Augmentation}

This project faced considerable data limitations due to the restricted availability of labeled foam images specifically collected from WTPs. To address this challenge, a variety of data strategies were employed, including a zero-shot approach, synthetic data generation, and manual segmentation.

\subsubsection{Water Dataset for Zero-Shot Prediction}

\begin{figure}[htp]
\centering
\includegraphics[width=0.6\textwidth]{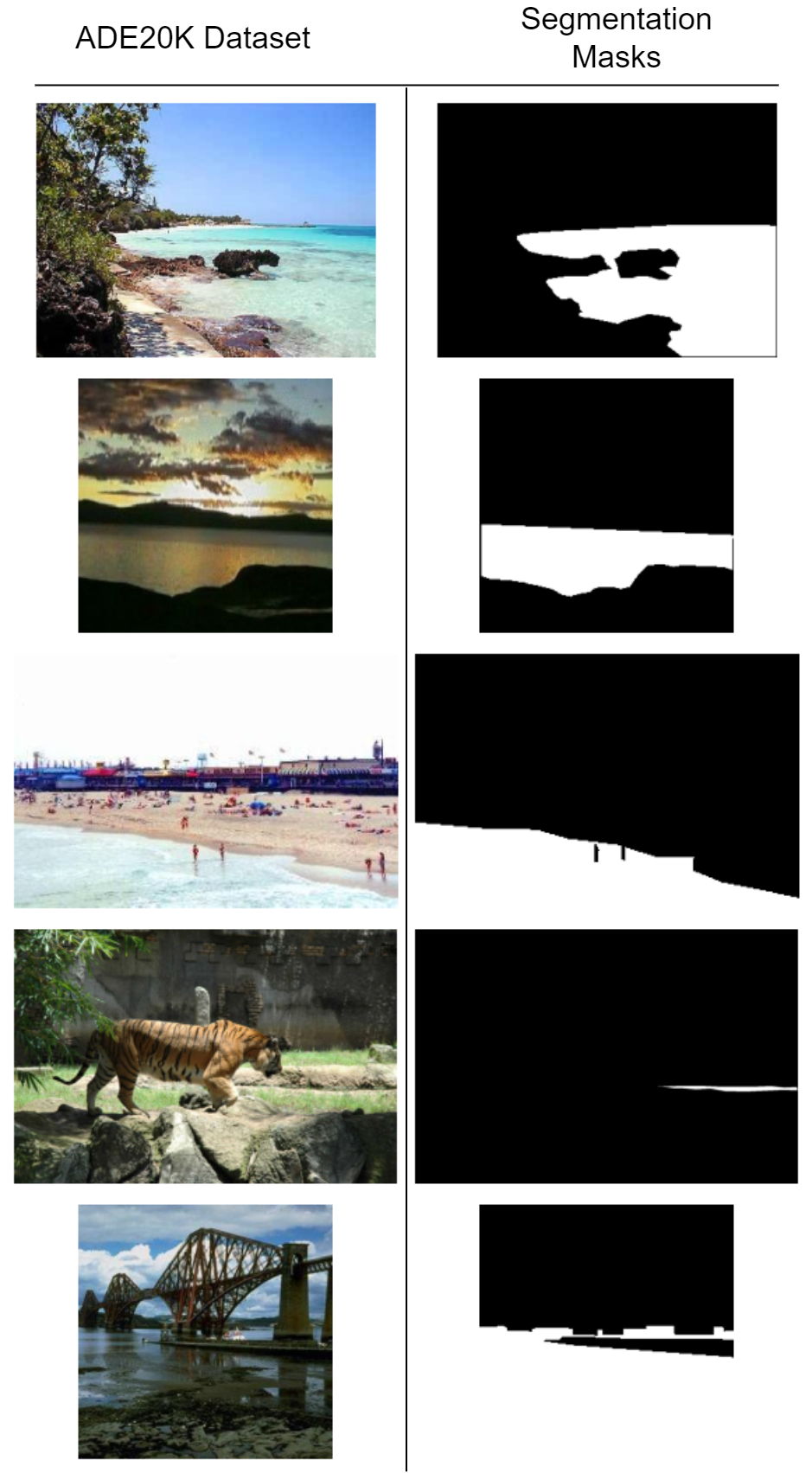}
\caption{Sample images from the ADE20K dataset used for zero-shot segmentation}
\label{fig:ade20k_examples}
\end{figure}

To overcome the initial scarcity of foam data, a zero-shot learning approach was adopted utilizing the ADE20K dataset \cite{liang2020waternet}, which serves as a benchmark for image segmentation and contains diverse visual information. After training on the ADE20K dataset, the model was tasked with classifying the remaining images, which were considered foam. This approach proved effective in the initial stages but led to segmentation errors, highlighting the need for more targeted datasets. In this dataset, only ADE20K data is used, with 3020 frames for training and 756 frames for testing. Example images from this dataset are shown in Figure \ref{fig:ade20k_examples}.

\subsubsection{Synthetic Data Set with Turbo SDXL Diffusion Model}

\begin{figure}[htp]
\centering
\includegraphics[width=0.6\textwidth]{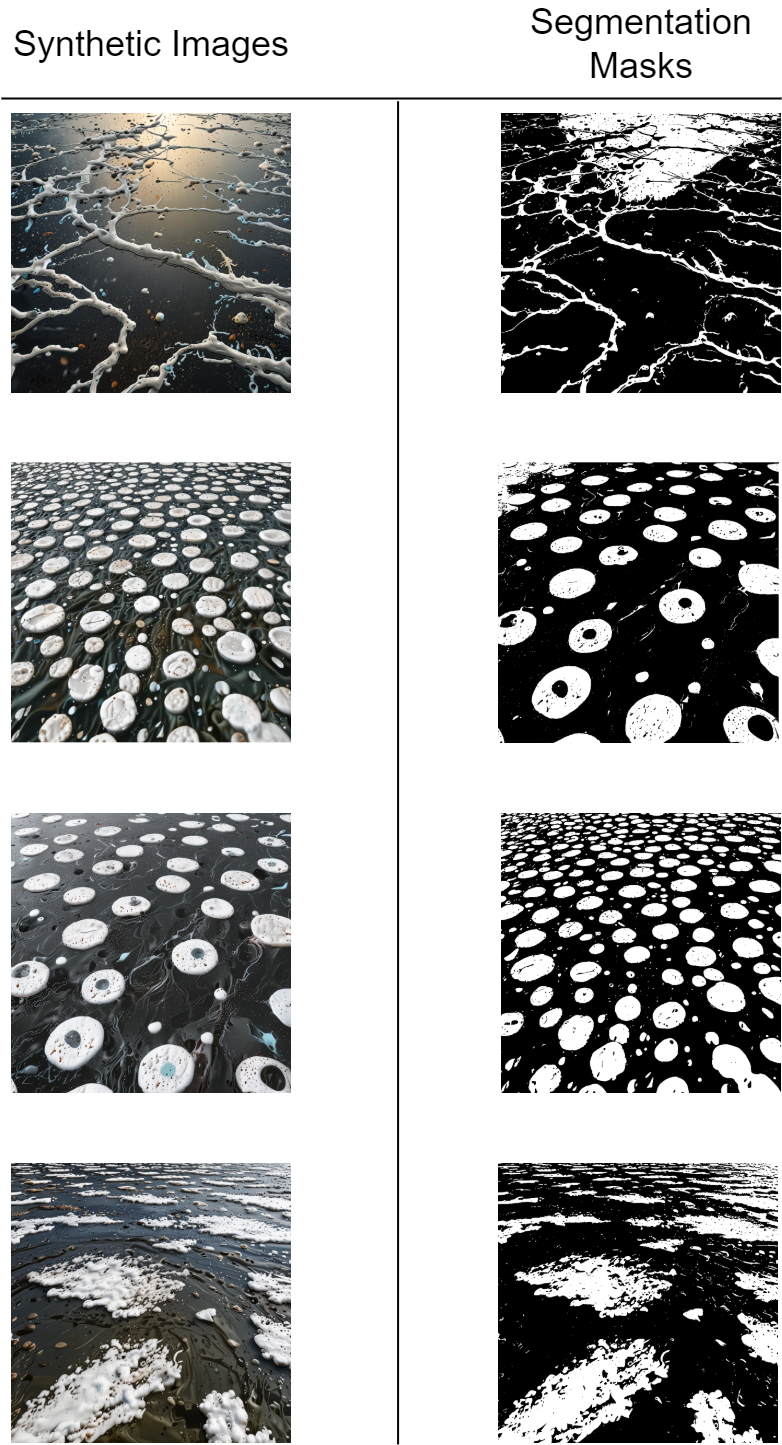}
\caption{Synthetic foam/water images generated via Turbo SDXL}
\label{fig:synthetic_examples}
\end{figure}

To further augment our datasets, synthetic foam and water images were generated using specific prompts within the Turbo SDXL diffusion model \cite{sauer2024adversarial}. However, this method remains experimental and its efficacy in this application requires further evaluation. The dataset contains 300 images. Representative examples are illustrated in Figure \ref{fig:synthetic_examples}

\subsubsection{Dataset from WTPs Images}
\label{sec:dataset_wwtp}

\begin{figure}[htp]
    \centering
    \includegraphics[width=0.9\linewidth]{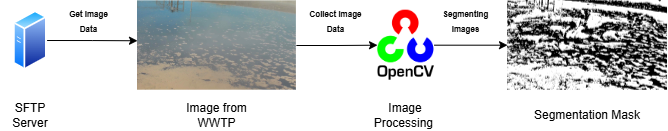}
    \caption{Overall workflow of day/night segmentation. The raw images are taken from the WTPs, pre-processed via OpenCV-based contrast adjustments (middle), and thresholded adaptively to yield a final foam mask}
    \label{fig:manuel_dataset_pipeline}
\end{figure}

To optimize model performance and capture foam variations over a typical 24-hour cycle, images were collected from a specific WTP located in Granada, Spain, at 10-minute intervals on selected days in October 2024. This dataset was captured within the context of the Zerovision project, in which the authors participated. The Zerovision projects aims to improve the wastewater management through Computer Vision and Federated AI. After excluding missing or low-quality images, a dataset consisting of 1,792 images suitable for analysis was compiled. These images were then processed by an automatic segmentation pipeline, outlined in Algorithm \ref{algo:segmentation}, to generate foam/no-foam binary masks for subsequent model training. The overall pipeline is visualized in Figure \ref{fig:manuel_dataset_pipeline}.

\begin{figure}[htp]
\centering
\includegraphics[width=0.6\textwidth]{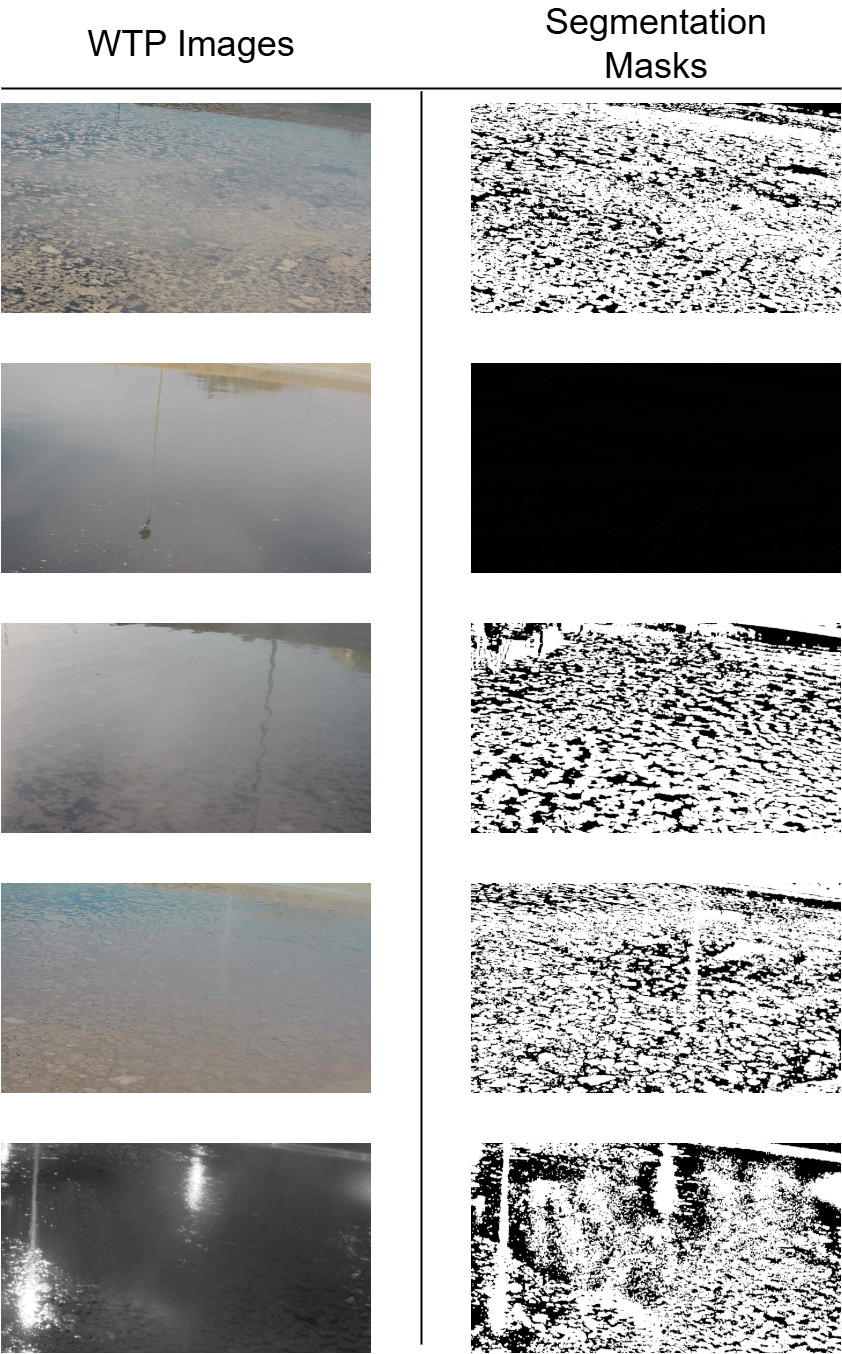}
\caption{Sample images from the real-world WTPs dataset collected in Granada}
\label{fig:wtp_examples}
\end{figure}

In this procedure, the \emph{day vs.\ night} branch (Step 2 in Algorithm \ref{algo:segmentation}) is determined by the average brightness in the grayscale image. Night images undergo additional brightening and denoising prior to thresholding, whereas day images rely on a CLAHE-based approach to handle typical lighting conditions. The morphological opening and connected-component filtering steps ensure that small, spurious regions do not appear in the final mask. This automatic labeling pipeline thus accelerates data preparation, allowing the dataset which contains 1,792 images to be transformed rapidly into foam/no-foam masks for subsequent model training and validation. Example real WTPs images are provided in Figure \ref{fig:wtp_examples}.

\begin{algorithm}[htp]
\caption{Automatic Day/Night Segmentation \& Mask Generation}
\label{algo:segmentation}
\begin{algorithmic}[1]

\Require \textit{Input directory} containing RGB images; \textit{Output directory} for segmented results
\Ensure Binary mask (\texttt{\_mask.jpg}) and overlay image (\texttt{\_overlay.jpg}) for each input

\For{each \textit{image} in \textit{input directory}}
    \State \textbf{Load and Convert:}
    \Statex \quad 1. Read the image from disk and convert to grayscale
    \Statex \quad 2. Compute the average brightness $b = \mathrm{mean}(\text{grayscale})$

    \If{$b < 100$ \textbf{(Night)}}
        \Statex \quad - Brighten \& enhance contrast via \texttt{convertScaleAbs} 
        \Statex \quad - Denoise using \texttt{fastNlMeansDenoising}
        \Statex \quad - Threshold using adaptive Gaussian
    \Else \textbf{(Day)}
        \Statex \quad - Use CLAHE for contrast-limited histogram equalization
        \Statex \quad - Threshold using adaptive Gaussian
    \EndIf

    \State \textbf{Morphology and Connected Components:}
    \Statex \quad - Apply morphological opening (3x3 kernel, 2 iterations) to remove noise
    \Statex \quad - Run connected-components analysis and discard regions with area $< 75$

    \State \textbf{Save Output:}
    \Statex \quad - Overlay the final mask onto the original RGB image, marking foam in color
    \Statex \quad - Write the binary mask (\texttt{\_mask.jpg}) and overlay (\texttt{\_overlay.jpg}) to the output directory
\EndFor
\end{algorithmic}
\end{algorithm}

\vspace{0.5em}
\noindent

\section{Methodology}
\label{sec:methodology}

This section describes how the methodology addresses critical challenges such as data privacy, inter-facility heterogeneity, and the need for robust segmentation under various operational conditions. The following subsections provide detailed information about our FL framework, the SAM2 adaptation, and data preprocessing techniques. Our solution consists of an open-source project\footnote{https://github.com/ertis-research/zerovision} made up of a series of components whose main objective is to offer a system for the autonomous detection of foam in secondary decanters in WTPs. The idea is to continuously perform complete sweeps in these water accumulations in search of foam formations, which are detrimental to the water cycle process due to the appearance of certain components. This is possible thanks to: 1) the use of our SAM2-based segmentation model for segmenting the foam on the water; 2) computing edge nodes in each of the decanters in which this model is deployed; 3) area scan cameras also located next to these nodes to obtain the images; and finally, 4) a Fog server that acts as the central axis and orchestrator of the inspections. In the latter, an FL process is carried out, fed by each of the worker (or child) nodes located in the decanters. In this way, a more complete and robust learning model is created with all the information coming from all the decanters, which also serves to update the segmentation models of all the nodes as the aggregation rounds pass. This architecture of parent-child nodes to perform FL can be scaled to work with different WTPs and thus increase the knowledge of our learning models. In other words, to carry out the desired task, the system comprises the following components:

\begin{itemize}[noitemsep]
    \item \textbf{Segmentation model}: SAM2-based segmentation model for the segmentation of foam in water.
    \item \textbf{Edge nodes}: Computing nodes where the segmentation models are deployed in order to carry out federated training in each of the decanters.
    \item \textbf{Area scan cameras}: Cameras that sweep the entire surface of the water in the decanters to obtain time-lapse images.
    \item \textbf{Fog server}: This is a central computing server that acts as the orchestrator for the Federated Learning process. Its primary role is to initiate training rounds, distribute the global model to the edge nodes, and aggregate the returning model weights (using the FedAvg algorithm) to create a new, improved global model. Critically, the Fog server never accesses the raw images, ensuring data privacy. In our deployment, this server was hosted on-premise to ensure data never left the facility.
\end{itemize}

\begin{figure}[htbp]
\centering
\includegraphics[width=0.8\textwidth]{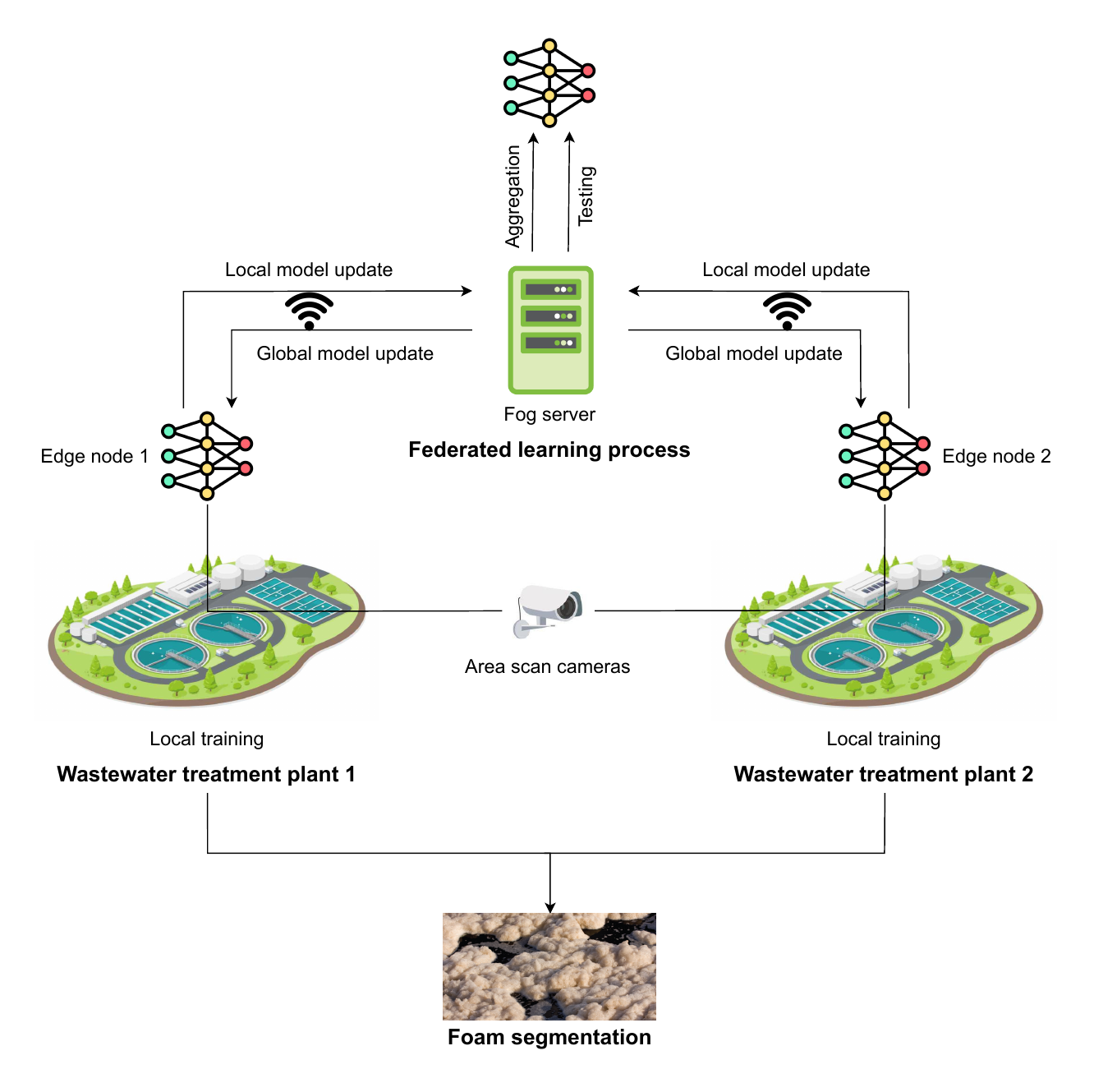} 
\caption{FL architecture for foam segmentation in WTPs}
\label{fig:architecture}
\end{figure}

Figure \ref{fig:architecture} shows the general FL architecture for foam segmentation in WTPs.

\subsection{Model Architecture: From SAM to SAM2}

The Segment Anything Model (SAM) \cite{kirillov2023segment} and its successor, SAM2 \cite{ravi2024sam}, are built on three core components:
\begin{itemize}[noitemsep]
    \item \textbf{Image Encoder}: A Vision Transformer (ViT) that converts the input image into a high-dimensional feature embedding.
    \item \textbf{Prompt Encoder}: Converts user inputs (such as points, bounding boxes, or text) into a separate embedding that specifies what to segment.
    \item \textbf{Mask Decoder}: Takes both the image and prompt embeddings and efficiently predicts the final segmentation mask.
\end{itemize}

While both models share this fundamental architecture, they differ significantly in their fine-tuning strategy and hierarchical feature extraction. As we discuss in Section \ref{sec:challenges}, our initial experiments with the original SAM model revealed limitations in fine-tuning for our specific task, leading us to discard it. We therefore adopted SAM2, which is designed for more flexible adaptation.

The following subsections detail how we fine-tuned these components for foam segmentation.

\begin{figure}[htp]
\centering
\includegraphics[width=0.6\textwidth]{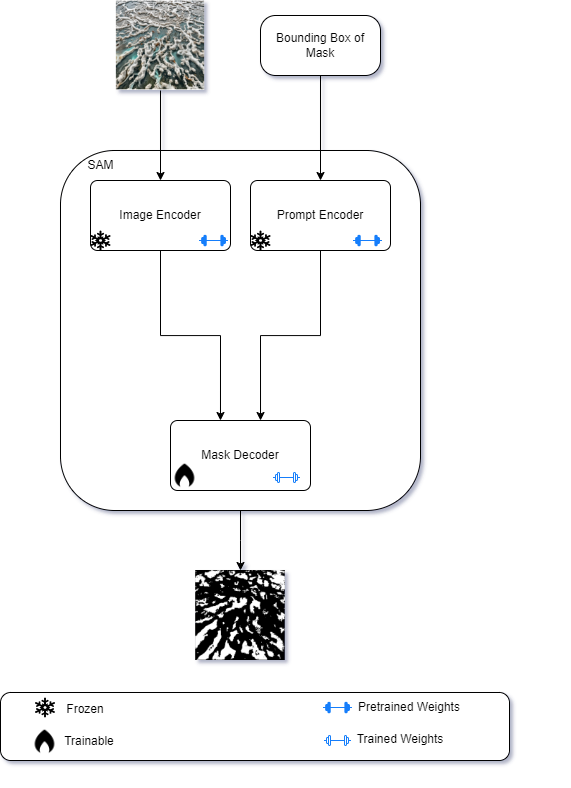} 
\caption{SAM model fine-tune architecture}
\label{fig:sam_training}
\end{figure}
\vspace{-1em} 

\begin{figure}[htbp]
\centering
\includegraphics[width=0.4\textwidth]{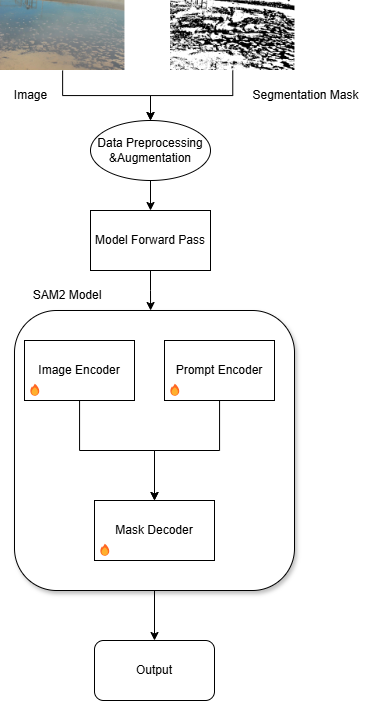} 
\caption{SAM2 model fine-tune architecture}
\label{fig:sam2_training}
\end{figure}

\subsubsection{Image Encoder}
\begin{description}[noitemsep, leftmargin=1em, style=nextline]
  \item[\textbf{SAM (v1)}]
  In the initial experiments utilizing the original SAM, the ViT-B variant was adopted as the image encoder; however, it was kept \emph{frozen} in order to preserve the rich and general-purpose features (see Figure \ref{fig:sam_training}). This was motivated by our relatively small foam dataset, where unfreezing the encoder could pose a risk of overfitting and destructive forgetting.

  \item[\textbf{SAM2}]
  In the subsequent approach employing SAM2, the image encoder was optionally fine-tuned to more effectively capture domain-specific nuances of foam images. SAM2 introduced a hierarchical vision transformer that extracted multi-scale feature maps, thereby refining foam boundaries essential for accurate segmentation (see Figure \ref{fig:sam2_training}). Depending on the size and complexity of the dataset, certain layers were still frozen to mitigate overfitting.
\end{description}

\subsubsection{Prompt Encoder}
\begin{description}[noitemsep, leftmargin=1em, style=nextline]
  \item[\textbf{SAM (v1)}]
  Originally, bounding box prompts were adapted for coarse localization, enabling SAM to refine the foam regions within these delineated areas.

  \item[\textbf{SAM2}]
    For training SAM2 within the FL framework, point prompts were automatically generated from the ground truth segmentation masks available in each client’s local dataset ($D_k$). Specifically, for each foam object instance in a mask, a random pixel coordinate belonging to that object was selected as a positive point prompt (label 1). For images containing no foam, a point near the image center with a negative label (0) was used as a negative prompt. While differing from interactive annotation, this strategy efficiently provides relevant spatial cues during fine-tuning by leveraging the available ground truth. This automated approach is suitable for the federated settings where manual interaction per client is impractical, ensuring the model learns to segment foam conditioned on receiving relevant point inputs during training.
\end{description}

\subsubsection{Mask Decoder}

\begin{description}[noitemsep, leftmargin=1em, style=nextline]
  \item[\textbf{SAM (v1)}]  
  The original decoder employed transformer blocks and CNN layers to produce segmentation masks. This decoder was primarily fine-tuned for the task of foam segmentation.

  \item[\textbf{SAM2}]  
  SAM2’s decoder includes additional multi-resolution features to more precisely delineate foam edges and shapes. The decoder continued to be fine-tuned, now augmented by the newly flexible encoder and the refined prompt encoder.
\end{description}

\paragraph{Loss Function.} 
Both SAM and SAM2 employ a combination of Dice or BCE losses for stable segmentation performance:

$$
\mathcal{L}_{\text{seg}} = \alpha \,\text{DiceLoss}(y, \hat{y}) + (1 - \alpha)\,\text{BCELoss}(y, \hat{y}),
$$
and SAM2 further adds a small penalty term aligning predicted mask-quality scores with Intersection over Union (IoU):

$$
\mathcal{L}_{\text{score}} = \left\lvert \text{score} - \text{IoU} \right\rvert.
$$
These modifications help ensure both accurate masks and well-calibrated confidence scores.

Overall, SAM2 proves well-suited to foam segmentation across heterogeneous wastewater treatment facilities. Our federated approach capitalizes on SAM2’s strong pre-trained features to achieve rapid convergence while safeguarding WTPs privacy, enabling robust foam detection even with limited or noisy local datasets.

\subsection{Federated Learning Framework}

The Flower Framework \cite{beutel2020flower} was utilized, a framework that has been demonstrated to be effective for FL and which integrates with libraries such as Pytorch and Tensorflow. A further rationale for our selection was the incorporation of FL algorithms, and the seamless integration with Pytorch. Flower also facilitates the modification of these FL algorithms. The implementation of an FL system is being conducted, with the FedAvg algorithm \cite{mcmahan2017communication} being utilized as the basis for this implementation. The FedAvg algorithm functions by initially transmitting the global model parameters from the server to a subset of clients. Subsequently, each client that has been selected proceeds to refine the model on its data for a number of epochs. Following this, the updated weights and, if desired, other parameters are dispatched back to the server. Subsequently, the server calculates the mean of the returned weights and creates a new global model. This process is repeated for the total number of rounds in each round. The design was modified to facilitate real-time monitoring of sent data, frequent logging, model evaluation on the server side, and detection of no changes in local training on the client side. This design enables collaborative model training without the need for centralized raw data. In light of the confidentiality of operational WTPs data, this is a critical requirement.

\vspace{0.5em}
\noindent
\textbf{FL Process Overview:} \\
The training process unfolds over several communication rounds, with each round comprising the following steps:

\begin{enumerate}
    \item \textbf{Initialization:} \\
    The server initializes the global model parameters $\theta_0$ using weights from a pre-trained SAM2 model (\textit{sam2\_hiera\_small.pt}) \cite{ravi2024sam}. This initialization leverages SAM2’s hierarchical vision transformer and robust multi-domain segmentation capabilities to accelerate convergence, especially valuable when clients possess limited real foam data initially.
    
    \item \textbf{Client Selection:} \\
    In each communication round $t$, the server selects a subset of WTPs clients for training using a round-robin strategy, ensuring that every site participates over time. Alternative selection policies (e.g., random or importance-based) may be adopted to optimize communication efficiency and maintain data diversity.

    \item \textbf{Local Training:} \\
    Each selected client $k$ downloads the current global model $\theta_t$ and fine-tunes it on its private dataset $D_k$, which may include manual labels, synthetic foam images, or newly generated WTPs masks. The training is performed with the AdamW optimizer \cite{loshchilov2017decoupled} using a learning rate $\eta = 1e-5$ and weight decay $\lambda = 4e-5$. Each client performs local fine-tuning for $30$ epochs, with each epoch comprising $5$ rounds using a batch size of $32$:
    
    $$
    \theta_{t+1,k} = \text{LocalTrain}(\theta_t, D_k, E, S, B, \eta, \lambda).
    $$

    AdamW’s robust handling of sparse gradients and large-scale parameters makes it particularly effective for segmentation tasks with hierarchical models like SAM2. In practice, a custom \texttt{FlowerClient} class manages parameter conversion (PyTorch \(\leftrightarrow\) NumPy) and delegates the training steps to a \texttt{SAM2Trainer}.

    \item \textbf{Model Aggregation:} \\
    After local training, clients upload their updated parameters $\theta_{t+1,k}$ to the server. The server aggregates these updates using a weighted average:
    
    $$
    \theta_{t+1} = \sum_{k} \Bigl(\frac{n_k}{n}\Bigr)\,\theta_{t+1,k},
    $$
    
    where $n_k$ is the number of samples in client $k$’s dataset and $n$ is the total number of samples across all clients. This weighted approach ensures that clients with larger or more diverse datasets (e.g., varying lighting conditions) have a proportionately greater influence on the global model. An \texttt{EnhancedFedAvg} strategy log fit metrics and updates the SAM2 model state accordingly.
    
    \item \textbf{Model Distribution:} \\
    Finally, the new global model $\theta_{t+1}$ is broadcast to all clients, replacing their local models and setting the stage for the next round of training.
\end{enumerate}
\noindent

The full pseudocode for the \emph{server} and \emph{client} logic is provided in Algorithm \ref{alg:fed_server} and Algorithm \ref{alg:flower_client}, respectively.

\begin{algorithm}[htp]
  \caption{Federated Server Main Loop}
  \label{alg:fed_server}
  \begin{algorithmic}[1]
    \Require  
      Number of rounds $R$, fraction clients $f_{\text{fit}},f_{\text{eval}}$,\\
      min clients $n_{\text{fit}},n_{\text{eval}},n_{\text{avail}}$,\\
      save directory $D_{\text{save}}$, model config $\mathcal{C}$, checkpoint $\theta_0$
    \Ensure  
      Aggregated model weights saved per round; global metrics logged

    \State Initialise logger
    \State Build initial SAM2 model $\theta \gets \textsc{BuildSAM2}(\mathcal{C},\theta_0)$
    \State Define strategy $S \gets \textsc{EnhancedFedAvg}(f_{\text{fit}},f_{\text{eval}},n_{\text{fit}},n_{\text{eval}},n_{\text{avail}},D_{\text{save}},\mathcal{C},\theta_0)$
    \State Start server: \texttt{fl.server.start\_server}(\textit{address}, $R$, $S$)
    
    \Function{EnhancedFedAvg.aggregate\_fit}{$r$, results, failures}
      \If{results empty} 
        \State Log warning; \Return none
      \EndIf
      \State Log aggregation start for round $r$
      \State $(\theta',\_) \gets \textsc{FedAvg.aggregate\_fit}(r,\,\dots)$
      \State Compute weighted metrics $\mu \gets \textsc{WeightedAverage}(\{\dots\})$
      \State Append $(r,\mu)$ to metrics lists
      \State \textsc{PlotMetrics}() and \textsc{LogCSV}()
      \State \textsc{SaveAggregatedModel}($r,\theta'$)
      \State \Return $(\theta',\mu)$
    \EndFunction

    \Function{EnhancedFedAvg.SaveAggregatedModel}{$r,\theta'$}
      \State Deserialize $\theta'$ into tensors
      \State Match names/shapes against $\theta$; build $\theta_{\text{agg}}$
      \State Load $\theta_{\text{agg}}$ into SAM2 model
      \State Save state to \texttt{D\_save/federated\_sam2\_round\_r.torch}
      \State Log success or warn on mismatches
    \EndFunction
  \end{algorithmic}
\end{algorithm}

\begin{algorithm}[htp]
  \caption{Federated Client Main Loop}
  \label{alg:flower_client}
  \begin{algorithmic}[1]
    \Require  
      Local trainer \textsc{SAM2Trainer}, iterations per round $I$
    \Ensure  
      Updated model parameters and local metrics

    \Function{get\_parameters}{}
      \State Extract $\theta$ from local SAM2 model
      \State \Return list of numpy arrays
    \EndFunction

    \Function{set\_parameters}{$\Theta$}
      \State Map $\Theta$ back into model state dict
      \State Load into SAM2 model
    \EndFunction

    \Function{fit}{$\Theta$, config}
      \State \Call{set\_parameters}{$\Theta$}
      \State Print “Starting local training”
      \State $(\ell, \mathrm{IoU}, \mathrm{Dice}, \mathrm{PA}) \gets$ \\
      \quad \Call{trainer.train}{epochs=30, steps=9, batch=32}
      \State Print “Training completed”
      \State \Return \Call{get\_parameters}{}, $n_{\text{samples}}$, \\
      \quad metrics \{loss:$\ell$, iou:$\mathrm{IoU}$, dice:$\mathrm{Dice}$, pixel\_accuracy:$\mathrm{PA}$\}
    \EndFunction

    \Function{evaluate}{$\Theta$, config}
      \State \Call{set\_parameters}{$\Theta$}
      \State Print “Starting evaluation”
      \State $(\ell, \mathrm{IoU}, \mathrm{Dice}, \mathrm{PA}) \gets$ \Call{trainer.evaluate}{samples=10}
      \State Print “Evaluation completed”
      \State \Return $\ell$, $n_{\text{samples}}$, \\
      \quad metrics as in \textsc{fit}
    \EndFunction

    \State \Call{fl.client.start\_numpy\_client}{server\_address, client=self}
  \end{algorithmic}
\end{algorithm}

For the experiments, five communication rounds were run on a relatively small number chosen based on early tests indicating swift convergence, likely due to SAM2’s strong initialization. In later rounds,     clients with more realistic foam masks, especially under various conditions (day/night), further improved segmentation performance in various WTPs environments. In the training part, two clients were used, one of which was trained with a dataset generated with images from a real WTPs in Granada, while the other was trained with a synthetic dataset.

\section{Implementation}
\label{sec:implementation}

The system workflow starts with area scan cameras capturing surface images from each decanter. These images are processed locally by edge nodes running an SAM2-based segmentation model to detect the presence of foam. Based on the analysis, the node records the results or triggers alerts if necessary. To ensure that the model is continuously adapted and improved without compromising data privacy, the Fog server periodically coordinates a FL process. This process collects information from all nodes to improve the global segmentation model. The FL part was covered in detail in the previous section, and this section covers how SAM2 was integrated into the WTP in Granada using the SFTP protocol and a PostgreSQL database.

The system implementation is structured into four main stages, each corresponding to a distinct pipeline component. This stages were adopted in the context of the ZeroVision project. For clarity, these stages are presented as separate subsections, each accompanied by a diagram illustrating the respective logic.

\subsection{System Hardware and Deployment Architecture}
\label{sec:hardware}
The framework is designed to integrate with existing WTP data infrastructure, which consists of three main components:

\begin{itemize}[noitemsep]
    \item \textbf{Image Source:} Pre-existing industrial CCTV cameras at the WTP. These cameras are part of an automated system that captures surface images and saves them to an on-premise SFTP server.
    \item \textbf{Edge Nodes (Clients):} An on-premise computing client (e.g., a local PC or server) that has access to the SFTP server. This node's role is to fetch the images, run the SAM2 model inference for foam detection, and during federated training securely fine-tune the model on its local data.
    \item \textbf{Fog Server (Orchestrator):} A central server that runs the Flower server-side logic. It orchestrates the FL rounds and aggregates model weights, never accessing the raw images.
\end{itemize}

This architecture, illustrated in the process flow in Figure \ref{fig:system_init}, ensures that raw data is processed locally and never leaves the facility, preserving data privacy. Our experimental setup, described in Section \ref{sec:evaluation}, simulates this client-server architecture.

\subsection{System Initialization}
\label{sec:system_init}

\begin{figure}[htp]
\centering
\includegraphics[width=\linewidth]{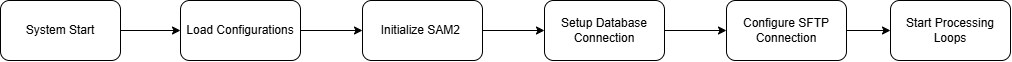}
\caption{System startup process workflow}
\label{fig:system_init}
\end{figure}

At startup, the system performs a series of initialization steps to prepare all components for operation (Figure \ref{fig:system_init}). First, it loads configuration parameters (such as camera settings, image capture frequency, processing thresholds, and SFTP server credentials) from a configuration file or environment variables. This ensures that all modules operate with the correct settings. Next, the hardware interfaces are initialized: for example, the camera or imaging sensor is activated and tested to confirm availability. In parallel, any required software resources are set up. This includes instantiating the image processing pipeline module and initializing data structures (buffers or queues) for inter-module communication. The system also spawns background threads or timers at this stage, such as a dedicated thread for periodic image acquisition and another for handling network communication. Finally, a self-check is performed to verify that every component is ready for normal operation. Only after successful verification does the system transition to the regular operational mode, beginning the main loop of acquisition and processing. If any critical failure occurs during initialization (for instance, a missing configuration file or an unavailable camera), the system logs the error and either halts startup or retries the initialization after a safe delay. This fail-safe behavior ensures robustness from the start by preventing the system from running in a compromised state.

\subsection{Image Acquisition Logic}

\subsection{Image Acquisition Logic}

\begin{figure}[htp]
\centering
\includegraphics[width=0.5\linewidth]{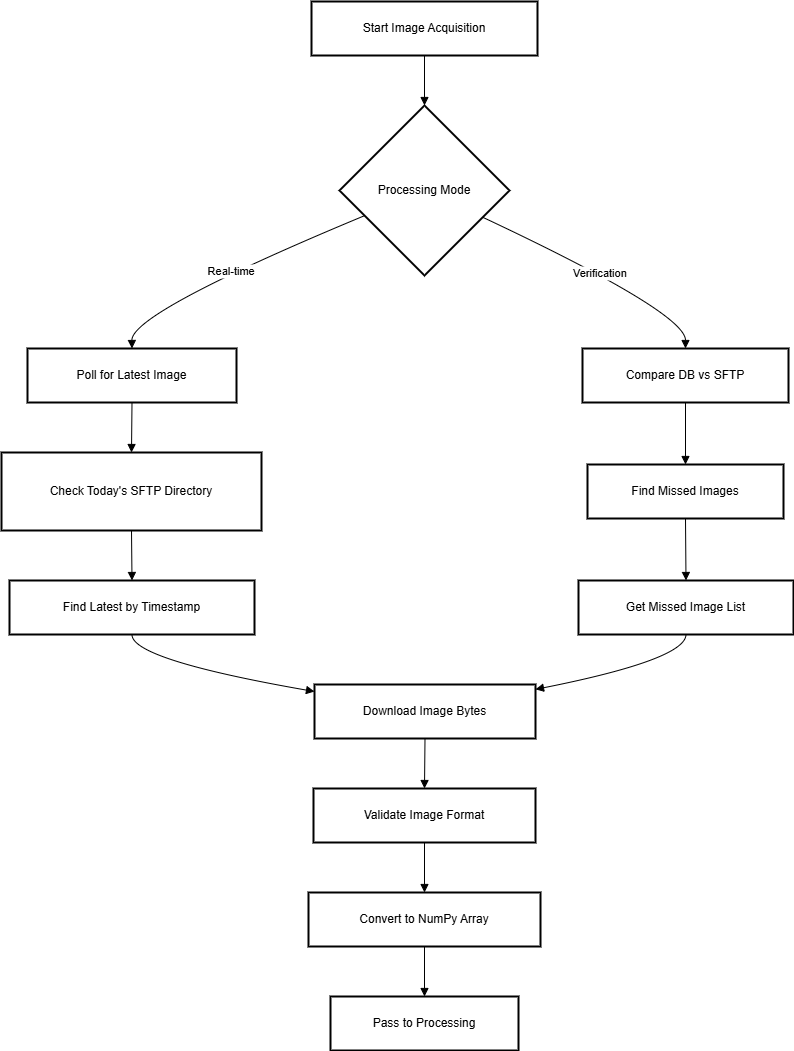}
\caption{Image acquisition workflow }
\label{fig:image_acquisition}
\end{figure}

The system does not process a live video feed. Instead, it fetches pre-captured images from the WTP's SFTP server, as detailed in Figure \ref{fig:image_acquisition}. This logic, which runs in a separate thread, operates in two modes:

\begin{itemize}[noitemsep]
    \item \textbf{Real-time Mode:} The primary loop for processing new data. The client polls the SFTP server by checking the directory for the current day and identifying the latest image by its timestamp.

    \item \textbf{Verification Mode:} A secondary process used to ensure no images were missed. It compares the list of processed images in the database against the full list of images on the SFTP server to find and retrieve any missed files.
\end{itemize}

In both modes, the selected image's bytes are downloaded, validated, and converted into a NumPy array. This array is then passed to the main processing pipeline. This SFTP-based, batch-processing approach is robust to the network and camera failures common in an industrial environment.

After all these processes, the processed image is sent to the model for processing, where SAM2 segments the image. The proportion of foamy areas in the predicted segmentation mask is sent to the database, and the image is saved to the SFTP server. All these operations proceed concurrently along with validation. Subsequently, after collecting a sufficient amount of data, the images are reviewed and converted into a dataset. Since each client's or WTPs's data is confidential, this data is trained locally at the plants. As shown in Figure \ref{fig:architecture}, after models are locally trained, they are sent to the Fog server where the models are aggregated, and the global model is updated. After the training process concludes, the global model is distributed back to the WTPs servers.

After all these processes, the processed image is sent to the model for processing, where SAM2 segments the image. The proportion of foamy areas in the predicted segmentation mask is sent to the database, and the image is saved to the SFTP server. All these operations proceed concurrently along with validation. Subsequently, after collecting a sufficient amount of data, the images are reviewed and converted into a dataset. Since each client's or WTPs's data is confidential, this data is trained locally at the plants. As shown in Figure \ref{fig:architecture}, after models are locally trained, they are sent to the Fog server where the models are aggregated, and the global model is updated. After the training process concludes, the global model is distributed back to the WTPs servers.

\section{Evaluation}
\label{sec:evaluation}

Experimental evaluations conducted on real-world WTPs data demonstrate the efficacy of our approach. FL initialized with SAM2 was observed to converge in significantly fewer communication rounds compared to random initialization. Furthermore, which represents more realistic scenarios, SAM2 initialization achieves a Dice score of 85.38\%, outperforming random initialization by a substantial margin (74.67\%).

Our experiments were conducted using PyTorch and the Flower framework to simulate the federated architecture described in Section \ref{sec:hardware}. All computations were performed on our research servers, each equipped with Intel Xeon Gold 6230R CPUs, NVIDIA Tesla V100 (32GB) GPUs, and 384 GB of RAM.

In this simulation, the Fog Server (FL server) and the Edge Node (FL client) were run as separate processes on this hardware. The client process was responsible for its local training, and the server process managed the aggregation, allowing us to accurately model the FL environment and measure performance.

Training times for the 75 epoch centralized baselines varied significantly. The U-Net and DeepLabV3+ models, using a ResNet34 backbone, both completed training in approximately 4 hours. In contrast, the SAM2-small model required considerably more computational time, finishing its 75 epochs in approximately 6.5 hours. This equates to an average of 5 minutes per epoch for the SAM2 model.

This epoch time has direct implications for federated training. In our FL experiment (which used the SAM2-BasePlus model), a single round with 30 local epochs required approximately 2.5 hours of training (30 epochs × 5 min/epoch) on each client. Since clients train in parallel, the total 5-round experiment took approximately 12.5 hours (5 rounds × 2.5 hours/round) to complete, plus negligible time for server-side aggregation.

\subsection{Evaluation Metrics}

The following metrics were used to evaluate the performance of the foam segmentation model:

\begin{enumerate}[noitemsep, label=\textbf{\arabic*.}]
    \item \textbf{Dice Coefficient.} Quantifies the overlap between the predicted mask and the ground truth, emphasizing balance between false positives and false negatives.
    \item \textbf{Intersection over Union (IoU).} Also known as the Jaccard index, measures the ratio of the intersection area to the union area of prediction and ground truth.
    \item \textbf{Pixel Accuracy.} Represents the proportion of correctly classified pixels across the entire image, offering an overall accuracy measure.
    \item \textbf{Loss.} Tracks both training and validation loss curves to monitor convergence and to detect potential overfitting.
\end{enumerate}

These metrics are computed both globally (across all clients) and locally (for each client) to assess the model's overall performance and its effectiveness in handling data heterogeneity. The mean and standard deviation of these metrics were reported across clients to provide a comprehensive view of model performance.

\subsection{Baseline Comparisons and Experimental Results}

To evaluate the effectiveness of our SAM-initialized FL approach, it was compared against the following baselines:

\begin{enumerate}[noitemsep, label=\textbf{\arabic*.}]
    \item \textbf{Centralized training of SAM2 and comparison with other segmentation models:} This baseline value represents the upper limit of performance that could be achieved if all data could be centralized (but this is not possible in practice due to privacy concerns). Variants of SAM2, Unet, and Deeplabv3+ were also compared in this section. All models in this section were trained using images and masks obtained from WTPs, as described in the dataset section.
    \item \textbf{FL with random initialization:} In this section, the model was trained in a federated manner. Two clients were used, one of which was trained with a dataset created from images obtained from WTPs, and the other with a synthetic dataset. This baseline helped quantify the benefits of using SAM2 as an initialization point for FL.
\end{enumerate}

\subsection{Local Training Hyperparameters}

The hyperparameters used in the local training phase of the models are as follows. For all variants of \textbf{SAM2}, learning rate = $1\times10^{-5}$, weight decay = $4\times10^{-5}$, loss = Dice loss, optimizer = AdamW, batch size = 32, and epochs = 75. For \textbf{U-Net}, the learning rate is $1\times10^{-3}$, batch size = 8, encoder = ResNet34 (pre-trained on ImageNet), loss = Dice loss, optimizer = Adam, and epochs = 75. For the \textbf{DeepLabV3+} model, the learning rate is $1\times10^{-3}$, encoder = ResNet34 (pre-trained on ImageNet), loss = BCEWithLogitsLoss, optimizer = Adam, and epochs = 75.

\subsection{Local Training Model Metrics Comparison}

\begin{figure}[htp] 
  \centering
  \includegraphics[width=0.8\linewidth]{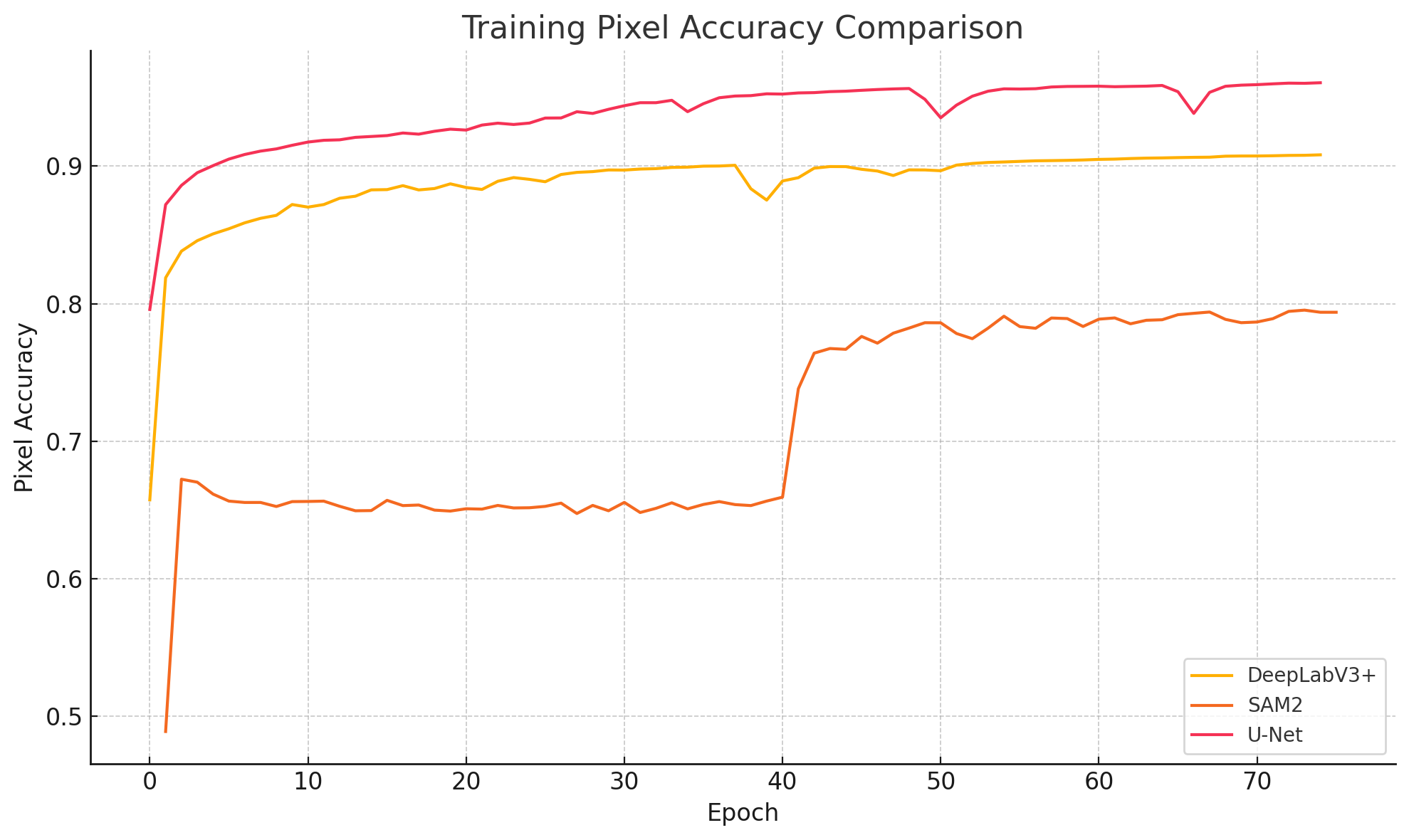}
  \caption{Pixel-accuracy trajectory for the locally trained baseline models}
  \label{fig:training-curves}
\end{figure}

\begin{figure}[htp]
  \centering
  \includegraphics[width=0.8\linewidth]{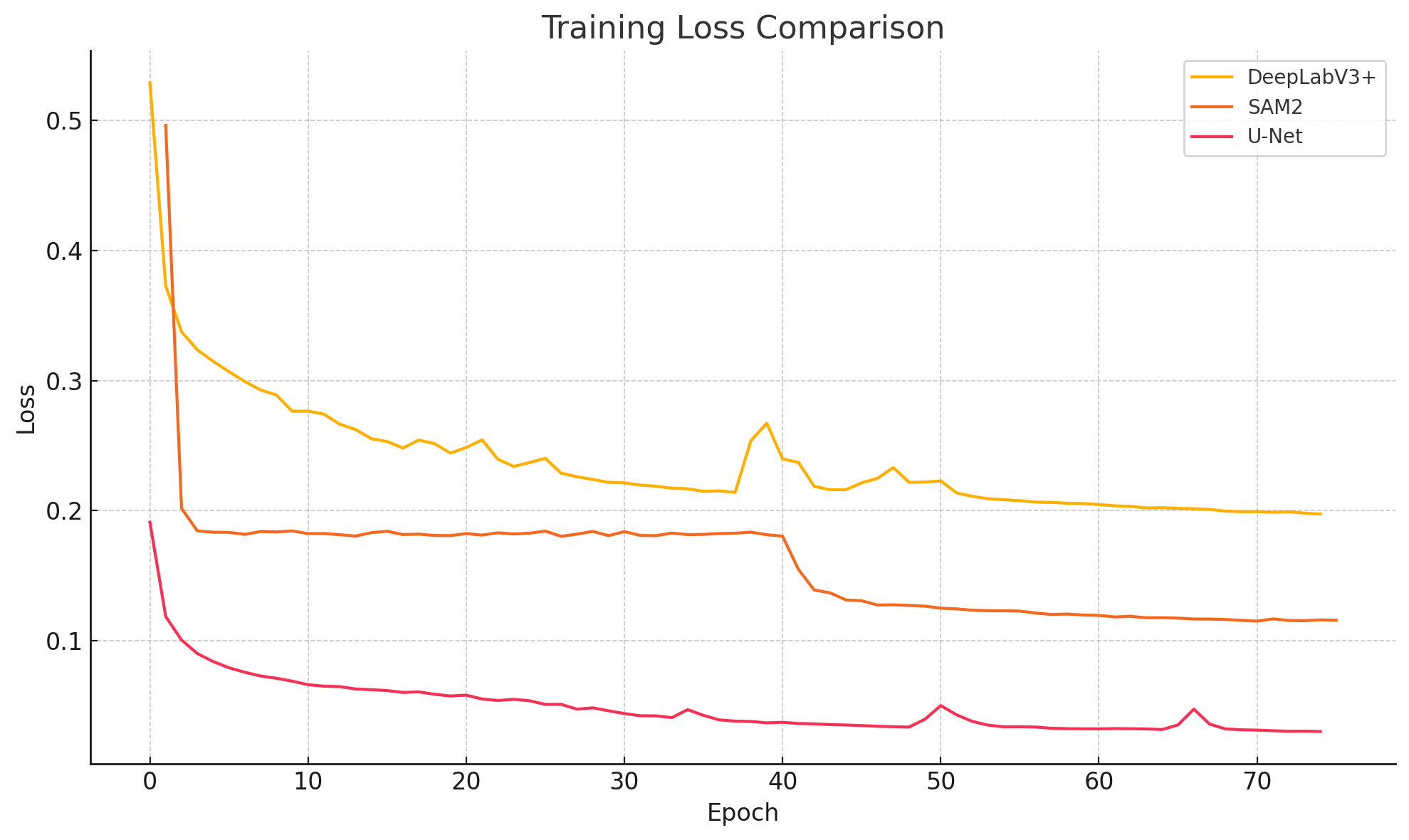}
  \caption{Corresponding training-loss curves}
  \label{fig:training-loss}
\end{figure}

As can be seen in Figures \ref{fig:training-curves} and \ref{fig:training-loss},  U-Net not only achieves the best final accuracy but also converges within ten epochs. DeepLabV3+ was shown to exhibit a slower yet steady improvement, whereas SAM2 stabilized around epoch 40 because, as mentioned in the Prompt-Sensitivity Analysis section, SAM2 could only segment images with excessive detail, such as foam, to a certain level due to the point prompt. Finally, the quantitative results are depicted in Table \ref{tab:train-final}.

\begin{table}[h]
  \centering
  \caption{Final training metrics normal training}
  \label{tab:train-final}
  \begin{tabular}{l
                  S[table-format=1.3]
                  S[table-format=1.3]
                  S[table-format=1.3]
                  S[table-format=1.3]}
    \toprule
    Model     & {Loss} & {Dice} & {IoU} & {Pixel Accuracy} \\
    \midrule
    DeepLabV3+ & 0.197  & 0.929  & 0.868 & 0.908            \\
    U-Net        & \textbf{0.030}  & \textbf{0.970}  & \textbf{0.942} & \textbf{0.961}            \\
    SAM2-tiny      & 0.112  & 0.844  & 0.758 & 0.797            \\
    SAM2-small     & 0.114  & 0.831  & 0.743 & 0.786            \\
    SAM2-baseplus  & 0.119  & 0.838  & 0.748 & 0.787            \\
    \bottomrule
  \end{tabular}
\end{table}

\subsection{Inference-Time Strategy and Qualitative Results}
\begin{figure}[htp]
    \centering
    \includegraphics[width=0.8\linewidth]{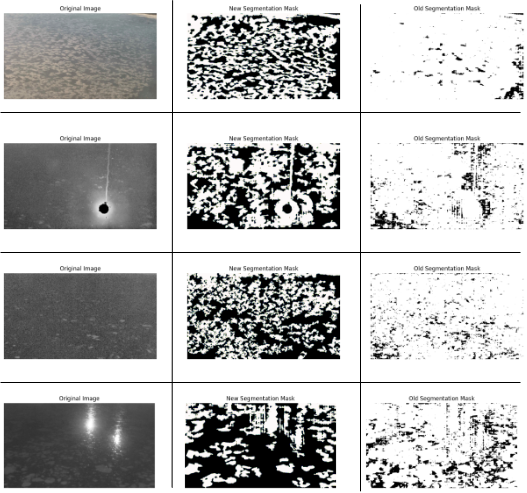}
    \caption{Comparison of segmentation results. The "New Segmentation Mask" column, generated using the inference algorithm, shows a marked improvement in detail and coverage over the "Old Segmentation Mask."}
    \label{fig:inference_comparison}
\end{figure}

During inference, a specialized processing pipeline was developed to enhance segmentation quality. This approach, detailed in Algorithm \ref{alg:inference}, significantly improves the model's ability to accurately identify foam  regions. It employs a systematic, grid-based prompting strategy combined with advanced denoising and mask refinement techniques. Figure \ref{fig:inference_comparison} illustrates that this new method produces substantially cleaner and more comprehensive segmentation masks than the older approach, capturing fine details and effectively reducing noise.

\begin{algorithm}
\caption{Inference-Time Segmentation Pipeline}\label{alg:inference}
\begin{algorithmic}[1]
\State \textbf{Input:} Raw image $I$, Number of points $N$
\State \textbf{Output:} Final segmentation mask $M_{final}$

\Procedure{SegmentFoam}{$I, N$}
    \State $I_{resized} \gets \text{ResizeImage}(I, \text{max\_dim}=1024)$
    \State $I_{denoised} \gets \text{BilateralFilter}(I_{resized})$
    \State $I_{denoised} \gets \text{FastNlMeansDenoising}(I_{denoised})$

    \State $P_{grid} \gets \text{GenerateGridPoints}(I_{denoised}, N)$
    \State Load pre-trained SAM2 model and initialize predictor

    \State Set predictor image to $I_{denoised}$
    \State $M_{raw}, S, L \gets \text{predictor.predict}(P_{grid})$

    \State $M_{refined} \gets \text{RefineMasks}(M_{raw}, S, \text{overlap\_threshold}=0.3)$
    
    \State $M_{binary} \gets (M_{refined} > 0) \times 255$
    \State $K \gets \text{GetStructuringElement}(\text{MORPH\_ELLIPSE})$
    \State $M_{opened} \gets \text{MorphologyEx}(M_{binary}, \text{MORPH\_OPEN}, K)$
    \State $M_{closed} \gets \text{MorphologyEx}(M_{opened}, \text{MORPH\_CLOSE}, K)$
    
    \State $M_{final} \gets \text{RemoveSmallComponents}(M_{closed}, \text{min\_area\_frac}=0.002)$
    
    \State \textbf{return} $M_{final}$
\EndProcedure
\end{algorithmic}
\end{algorithm}

\subsection{Qualitative Evaluation}

\begin{figure}[htp]
  \centering
  \includegraphics[width=0.8\linewidth]{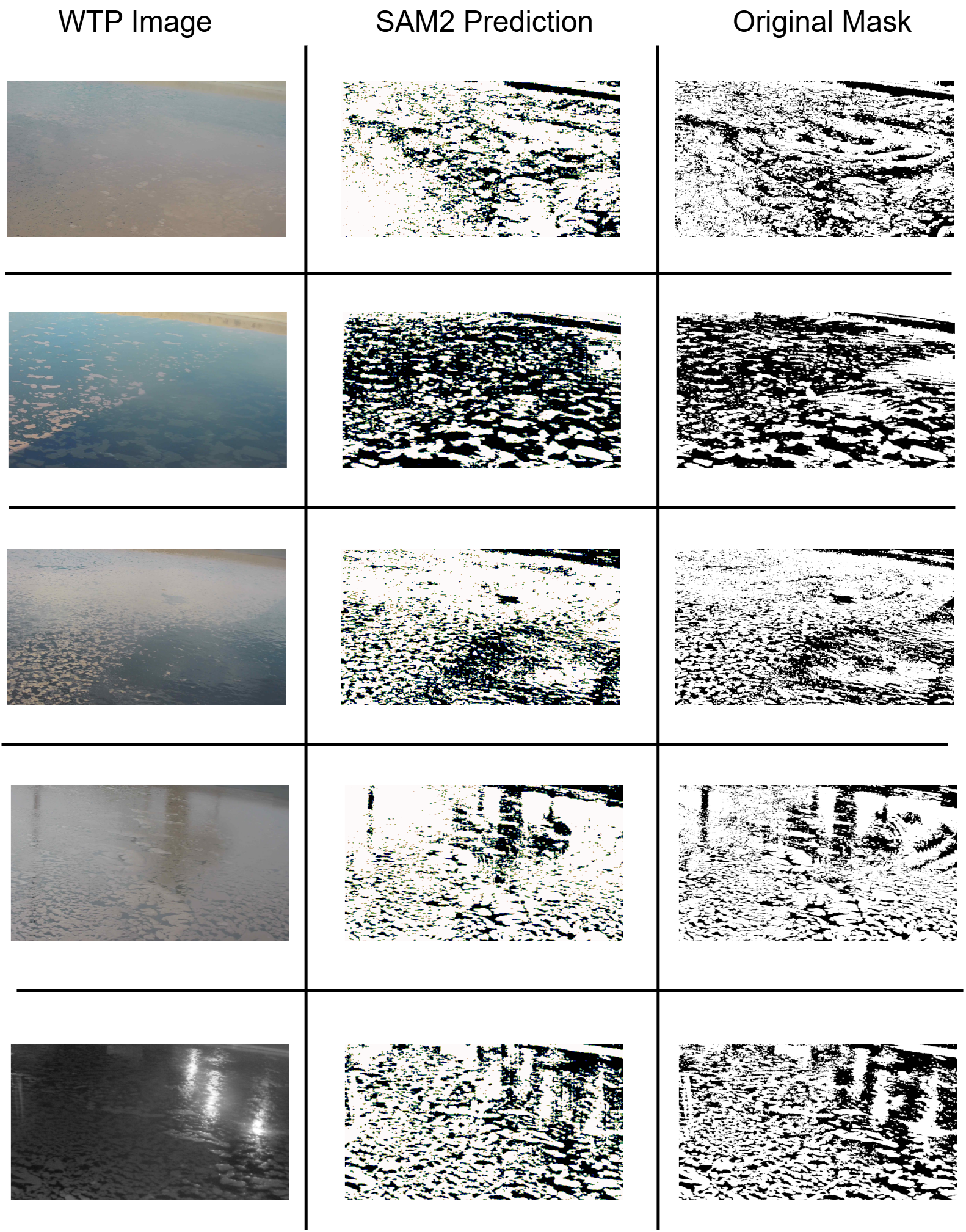}
  \caption{Example comparison between original images, ground truth masks, and SAM2 model predictions from the Granada WTPs dataset}
  \label{fig:sam2_example_results}
\end{figure}

To illustrate the segmentation capability of the trained SAM2 model, Figure \ref{fig:sam2_example_results} presents visual examples comparing the original masks and predicted segmentation masks.

\subsection{Federated Training of SAM2}

In this section, two federated clients (one for each sedimentation tank) were trained for 5 rounds, each with 30 epochs, using the FedAvg algorithm. In one model, the dataset created from images obtained from the water treatment plant (WTPs) was used, while in the other, the synthetic dataset generated using Turbo SDXL was utilized. The parameters used, with the hyperparameters of the two clients set equal to each other, were as follows: learning rate = $1\times10^{-5}$, weight decay = $4\times10^{-5}$, loss = Dice loss, optimizer = AdamW, batch size = 32, and epochs = 75.

\begin{table}[htp]
  \centering
  \caption{Aggregated metrics for \textbf{SAM2-BasePlus} during federated learning}
  \label{tab:fl-sam2-baseplus}
  \begin{tabular}{
    S[table-format=1.0]
    S[table-format=1.3]
    S[table-format=1.3]
    S[table-format=1.3]
    S[table-format=1.3]}
    \toprule
    {Round} & {Loss} & {IoU} & {Pixel Accuracy} & {Dice} \\
    \midrule
    1 & 0.139 & 0.727 & 0.778 & 0.825 \\
    2 & 0.123 & 0.749 & 0.802 & 0.841 \\
    3 & 0.118 & 0.755 & 0.807 & 0.844 \\
    4 & 0.115 & 0.758 & 0.809 & 0.845 \\
    5 & 0.113 & 0.762 & 0.814 & 0.848 \\
    \bottomrule
  \end{tabular}
\end{table}

\begin{figure}[htp] 
  \centering
  \includegraphics[width=0.8\linewidth]{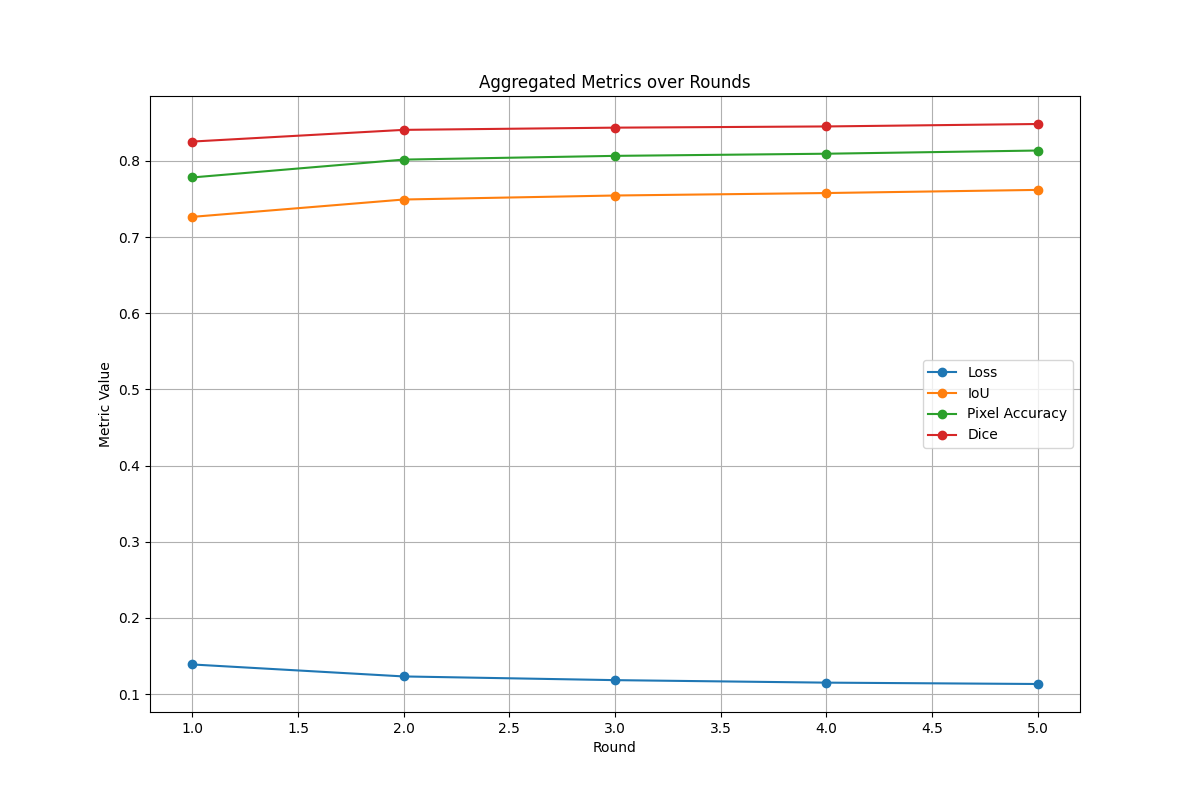}
  \caption{Aggregated metrics trajectory for \textbf{SAM2-BasePlus} during federated learning}
  \label{fig:federated-training-curves}
\end{figure}

When benchmarked against classical training, as shown by the results of every round in Table \ref{tab:train-final} and Figure \ref{fig:federated-training-curves}, the improvement observed under the federated setting could be partly attributed to the larger effective training budget: five communication rounds with 30\,local epochs each amount to roughly 150\,epochs in total. Nevertheless, the vanilla \emph{Federated Averaging} (FedAvg) algorithm remains a surprisingly strong baseline even in the presence of heterogeneous client data and provides a solid launch pad for more sophisticated FL strategies. Concretely, SAM2-BasePlus gains about \textbf{+2\,pp} in pixel-accuracy, Dice, and IoU over its single-site counterpart.

By providing this comprehensive methodology, the aim was to ensure the reproducibility of the results and to facilitate further research in the critical area of wastewater treatment plant monitoring and management using advanced machine learning techniques. Our approach represents a novel intersection of FL, transfer learning from foundation models, and domain-specific adaptation for environmental monitoring applications.

 \subsection{Prompt-Sensitivity Analysis} \label{prompt-sensitivity}
 
Since the SAM2 architecture is prompt-based (Point Input, Box Input, Prompt Input), it generally performs better on selected objects. However, it struggles to achieve the desired performance on objects with extremely fine details, such as foam, due to this structure. Nevertheless, as seen in general evaluations, the segmentation results on sample outputs and new images from the plant are almost identical to those of models like UNET and DeepLabV3+.

As shown in the training algorithm in the methodology section of the model's training, a dummy point prompt is created (randomly), and it is determined whether there is a foam mask in the index annotation. Based on whether the object exists, a positive or negative value is assigned. Therefore, the training phase yields lower pixel accuracy and IoU scores compared to other segmentation models. However, as seen in Table \ref{tab:train-final}, the Dice score is comparable to that of other models.

There are also efforts to improve this situation. As seen in the article by \cite{zhaksylyk2025rp}, it offers flexible ways to identify objects through text, mask, bounding box, or point prompts. However, point prompts inherently carry a flaw: even a slight change in the point’s location can significantly alter the model’s output. While most current models based on SAM2 focus on improving segmentation performance using high-quality prompts, segmentation consistency is limited when point prompts come from different regions of the object. This increases dependency on precise point placement by the annotator.

In RP-SAM2, a new Shift Block has been integrated, which relocates the user's input point prompt using cross-attention with image embeddings. During the fine-tuning process, the shift block was fine-tuned on 5–60\% of the training set and pseudo masks were generated for the remaining 40–95\%. Therefore, RP-SAM2 plays a significant role in fine-tuning by feeding pseudo-labels directly to a component of SAM2 (the mask encoder) to produce more accurate segmentation masks.

\section{Lessons learned}
\label{sec:challenges}

During the development of the model and the preparation of the dataset, various challenges were encountered. The first segmentation model employed was the SAM model. However, due to the lack of specific documentation and support for fine-tuning in the source code shared by Meta, Huggingface's Transformer Framework \cite{wolf-etal-2020-transformers} was utilized for fine-tuning. Despite our efforts, both the model and the dataset did not perform well at this stage. Additionally, it was realized that the Transformer library did not fully align with the Flower Framework, which was being used for federated learning. Consequently, the search for a different model and dataset was initiated.

While our work was ongoing, the SAM2 model was released. Along with fine-tuning support, we were also able to obtain the necessary images from a specific WTP, allowing us to build the dataset. We confirmed that SAM2 provided faster and better results compared to the first model. However, while constructing the dataset, we noticed that the glitter and reflections on the water did not yield good results with our image processing algorithm. As a result, we manually corrected certain areas. In some mask images without foam, we used black images. Despite CLAHE-based pre-processing and morphological cleaning, highly specular glints occasionally oversaturate the camera sensor, leading to false foam positives. However, during the inference phase after training the model, we observed that in some predictions, reflections were still being perceived as foam.

In addition to these, as mentioned in Section \ref{prompt-sensitivity}, prompts and point prompts play an important role in SAM2 training, so we could not get the same results as other segmentation models in our scenario during model training. We can partially solve the problem we could not solve in the training section in the implementation section. By thinking of this in terms of a formal grid, we can achieve good segmentation by specifying a certain number of notes (50 or more). However, as the number of points increases, the model's inference time also increases, so it is better to keep this at an optimal number.

\section{Conclusions and future work}
\label{sec:conclusion}

In this paper, we demonstrated the effectiveness of a solution for segmenting foam in wastewater treatment facilities. Notably, we introduced a practical FL pipeline designed to preserve data privacy while enabling collaborative model training. Regarding the models used, while SAM2's lightweight architecture and hierarchical transformer layers make it more adaptable for fine-tuning than its SAM predecessor, it does not always match the accuracy of task-specific networks like U-Net or DeepLabV3+ during fine-tuning on binary segmentation. However, experimental findings demonstrate that SAM2 attains superior generalization when deployed across heterogeneous plant data, a consequence of its pre-training on an unparalleled 11 million images and 11 billion masks. This substantial foundation endows SAM2 with more sophisticated visual priors in comparison to conventional backbones (VGG16, ResNet variants, DenseNet121, etc.), thereby facilitating its ability to sustain robust performance even under non-IID federated conditions.

The methodology included a comprehensive data strategy that combined real-world images from a WTPs in Granada (Spain), a synthetically generated dataset covering boundary conditions, and a publicly available dataset to improve generalization. We also added an automated mask generation pipeline to streamline the data preparation process. Through systematic evaluation, we demonstrated that centralized models, such as U-Net, can achieve higher metrics on a single local dataset, but our SAM2-based federated approach delivers robust and competitive performance under the practical constraints of a distributed, privacy-preserving system.

In summary, this research represents an important step in the intelligent monitoring of WTPs. The proposed framework not only provides a concrete solution for foam management, but also serves as a model for applying advanced, privacy-preserving machine learning techniques to other environmental and industrial monitoring challenges.

As future work, we have pointed out the following challenges and improvements to the proposed solution:

\begin{itemize}
    \item \textbf{Advanced Federated Learning Strategies}. While the FedAvg algorithm provided a strong baseline, future work could explore more sophisticated FL algorithms designed to better handle statistical heterogeneity across clients. Techniques such as FedProx, FedNova, or personalized FL could improve model performance and convergence speed by accounting for the unique data distributions at each WTPs.

    \item \textbf{Enhanced Prompt Engineering for SAM2}. As noted in Section \ref{prompt-sensitivity}, SAM2's performance is closely tied to the quality of its prompts. Future research could investigate more advanced methods for automatically generating high-quality point or box prompts during federated training. Furthermore, exploring architectures like RP-SAM2, which are designed to be more robust to prompt variations, could lead to more consistent and accurate segmentation of fine-detailed objects, such as foam. Experimenting with multi-point grids during inference, while balancing accuracy and computational cost, is another area for optimization.

    \item \textbf{Optimizing for Edge Deployment}. While the current implementation utilizes edge nodes, further optimization of the SAM2 model for resource-constrained edge devices is a valuable research direction. Techniques such as model quantization, knowledge distillation from the larger global model to a smaller, faster local model, and pruning could reduce the computational footprint and inference time, making real-time, on-site analysis more efficient.

    \item \textbf{Richer and fairer datasets}. Partner with multiple WTPs across climates to capture seasonal and process-driven variability. Expand synthetic data generation with physics-aware diffusion prompts that encode surfactant composition, aeration rates, and illumination spectra.
\end{itemize}

\section*{Acknowledgments}
 
This work is funded by the Spanish projects Grant CPP2021-009032 (`ZeroVision: Enabling Zero impact wastewater treatment through Computer Vision and Federated AI') funded by MICIU/AEI/10.13039/501100011033/ and by `European Union NextGenerationEU/PRTR', Grant CPP2022-009695 (`TRIATHLON: TRIhAlomeTHanes controL thrOugh AI-based techNologies' funded by MICIU/AEI/10.13039/501100011033/ and by `European Union NextGenerationEU/PRTR', and Grant PID2022-141705OB-C21 (`DiTaS: A framework for agnostic compositional and cognitive digital twin services') funded by MICIU/AEI/10.13039/501100011033/ and by `FEDER'.

\bibliographystyle{elsarticle-harv} 
\bibliography{references}

@Article{collivignarelli2020foams,
  title={Foams in wastewater treatment plants: from causes to control methods},
  author={Collivignarelli, Maria Cristina and Baldi, Marco and Abb{\`a}, Alessandro and Caccamo, Francesca Maria and Carnevale Miino, Marco and Rada, Elena Cristina and Torretta, Vincenzo},
  journal={Applied Sciences},
  volume={10},
  number={8},
  pages={2716},
  year={2020},
  publisher={MDPI}
}

@Article{ruzicka2009cause,
  title={Cause and effect relationship between foam formation and treated wastewater effluents in a transboundary river},
  author={Ruzicka, Katerina and Gabriel, Oliver and Bletterie, Ulrike and Winkler, Stefan and Zessner, Matthias},
  journal={Physics and Chemistry of the Earth, Parts A/B/C},
  volume={34},
  number={8-9},
  pages={565--573},
  year={2009},
  publisher={Elsevier}
}

@Article{wang2013enhancing,
  title={Enhancing the adsorption of the proteins in the soy whey wastewater using foam separation column fitted with internal baffles},
  author={Wang, Lianjie and Wu, Zhaoliang and Zhao, Bin and Liu, Wei and Gao, Yanfei},
  journal={Journal of Food Engineering},
  volume={119},
  number={2},
  pages={377--384},
  year={2013},
  publisher={Elsevier}
}

@Article{Karakashev2012,
  author = {Karakashev, D. and Grozdanova, M.},
  title = {Physicochemical properties of foam in wastewater treatment plants},
  journal = {Desalination and Water Treatment},
  volume = {38},
  number = {1-3},
  pages = {253-261},
  year = {2012}
}

@inproceedings{ronneberger2015u,
  title={U-net: Convolutional networks for biomedical image segmentation},
  author={Ronneberger, Olaf and Fischer, Philipp and Brox, Thomas},
  booktitle={Medical image computing and computer-assisted intervention--MICCAI 2015: 18th international conference, Munich, Germany, October 5-9, 2015, proceedings, part III 18},
  pages={234--241},
  year={2015},
  organization={Springer}
}

@inproceedings{long2015fully,
  title={Fully convolutional networks for semantic segmentation},
  author={Long, Jonathan and Shelhamer, Evan and Darrell, Trevor},
  booktitle={Proceedings of the IEEE conference on computer vision and pattern recognition},
  pages={3431--3440},
  year={2015}
}

@Article{carballo2024foam,
  title={Foam Segmentation in Wastewater Treatment Plants},
  author={Carballo Mato, Joaqu{\'\i}n and Gonz{\'a}lez V{\'a}zquez, Sonia and Fern{\'a}ndez {\'A}guila, Jes{\'u}s and Delgado Rodr{\'\i}guez, {\'A}ngel and Lin, Xin and Garabato G{\'a}ndara, Luc{\'\i}a and Sobreira Seoane, Juan and Silva Castro, Jose},
  journal={Water},
  volume={16},
  number={3},
  pages={390},
  year={2024},
  publisher={Multidisciplinary Digital Publishing Institute}
}

@inproceedings{mcmahan2017communication,
  title={Communication-efficient learning of deep networks from decentralized data},
  author={McMahan, Brendan and Moore, Eider and Ramage, Daniel and Hampson, Seth and y Arcas, Blaise Aguera},
  booktitle={Artificial intelligence and statistics},
  pages={1273--1282},
  year={2017},
  organization={PMLR}
}

@inproceedings{kirillov2023segment,
  title={Segment anything},
  author={Kirillov, Alexander and Mintun, Eric and Ravi, Nikhila and Mao, Hanzi and Rolland, Chloe and Gustafson, Laura and Xiao, Tete and Whitehead, Spencer and Berg, Alexander C and Lo, Wan-Yen and others},
  booktitle={Proceedings of the IEEE/CVF International Conference on Computer Vision},
  pages={4015--4026},
  year={2023}
}

@Article{loshchilov2017decoupled,
  title={Decoupled weight decay regularization},
  author={Loshchilov, Ilya and Hutter, Frank},
  journal={arXiv preprint arXiv:1711.05101},
  year={2017}
}

@Article{raghu2019transfusion,
  title={Transfusion: Understanding transfer learning for medical imaging},
  author={Raghu, Maithra and Zhang, Chiyuan and Kleinberg, Jon and Bengio, Samy},
  journal={Advances in neural information processing systems},
  volume={32},
  year={2019}
}

@Article{Zhang2020,
  author = {Zhang, W. and Wang, F. and Zhou, X. and Zhu, Y. and Zhu, Q. and Zhang, H.},
  title = {A survey on federated learning},
  journal = {arXiv preprint arXiv:2008.07317},
  year = {2020}
}

@Article{newhart2019data,
  title={Data-driven performance analyses of wastewater treatment plants: A review},
  author={Newhart, Kathryn B and Holloway, Ryan W and Hering, Amanda S and Cath, Tzahi Y},
  journal={Water research},
  volume={157},
  pages={498--513},
  year={2019},
  publisher={Elsevier}
}

@Article{petrovski2011examination,
  title={An examination of the mechanisms for stable foam formation in activated sludge systems},
  author={Petrovski, Steve and Dyson, Zoe A and Quill, Eben S and McIlroy, Simon J and Tillett, Daniel and Seviour, Robert J},
  journal={Water research},
  volume={45},
  number={5},
  pages={2146--2154},
  year={2011},
  publisher={Elsevier}
}

@Article{batinovic2021cocultivation,
  title={Cocultivation of an ultrasmall environmental parasitic bacterium with lytic ability against bacteria associated with wastewater foams},
  author={Batinovic, Steven and Rose, Jayson JA and Ratcliffe, Julian and Seviour, Robert J and Petrovski, Steve},
  journal={Nature Microbiology},
  volume={6},
  number={6},
  pages={703--711},
  year={2021},
  publisher={Nature Publishing Group UK London}
}

@Article{driver2024encrypted,
  title={Encrypted data-sharing for preserving privacy in wastewater-based epidemiology},
  author={Driver, Erin M and Ahsan, Manazir and Piske, Lucas and Lee, Heewook and Forrest, Stephanie and Halden, Rolf U and Trieu, Ni},
  journal={Science of The Total Environment},
  volume={940},
  pages={173315},
  year={2024},
  publisher={Elsevier}
}

@Article{ravi2024sam,
  title={Sam 2: Segment anything in images and videos},
  author={Ravi, Nikhila and Gabeur, Valentin and Hu, Yuan-Ting and Hu, Ronghang and Ryali, Chaitanya and Ma, Tengyu and Khedr, Haitham and R{\"a}dle, Roman and Rolland, Chloe and Gustafson, Laura and others},
  journal={arXiv preprint arXiv:2408.00714},
  year={2024}
}

@inproceedings{subedi2023client,
  title={A client-server deep federated learning for cross-domain surgical image segmentation},
  author={Subedi, Ronast and Gaire, Rebati Raman and Ali, Sharib and Nguyen, Anh and Stoyanov, Danail and Bhattarai, Binod},
  booktitle={MICCAI Workshop on Data Engineering in Medical Imaging},
  pages={21--33},
  year={2023},
  organization={Springer}
}

@Article{ahmadi2024federated,
  title={Federated Learning Approach to Mitigate Water Wastage},
  author={Ahmadi, Sina Hajer and Mahashabde, Amruta Pranadika},
  journal={arXiv preprint arXiv:2409.03776},
  year={2024}
}

@Article{mehrnia2024assessing,
  title={Assessing Foundational Medical'Segment Anything'(Med-SAM1, Med-SAM2) Deep Learning Models for Left Atrial Segmentation in 3D LGE MRI},
  author={Mehrnia, Mehri and Elbayumi, Mohamed and Elbaz, Mohammed SM},
  journal={arXiv preprint arXiv:2411.05963},
  year={2024}
}

@Article{zhou2024adaptive,
  title={Adaptive segmentation enhanced asynchronous federated learning for sustainable intelligent transportation systems},
  author={Zhou, Xiaokang and Liang, Wei and Kawai, Akira and Fueda, Kaoru and She, Jinhua and Kevin, I and Wang, Kai},
  journal={IEEE Transactions on Intelligent Transportation Systems},
  year={2024},
  publisher={IEEE}
}

@inproceedings{li2022detection,
  title={Detection method of nickel foam foreign matter based on double-branch lightweight network},
  author={Li, Jianqi and Liang, Yincong and Liu, Huiting and Cao, Binfang and Yan, Wei and Du, Rui},
  booktitle={2022 China Automation Congress (CAC)},
  pages={4077--4080},
  year={2022},
  organization={IEEE}
}

@inproceedings{singh2024enhanced,
  title={Enhanced Pothole Detection Using YOLOv5 and Federated Learning},
  author={Singh, Aditya and Mehta, Aryan and Padaria, Ali Asgar and Jadav, Nilesh Kumar and Geddam, Rebakah and Tanwar, Sudeep},
  booktitle={2024 14th International Conference on Cloud Computing, Data Science \& Engineering (Confluence)},
  pages={549--554},
  year={2024},
  organization={IEEE}
}

@Article{sun2024fkd,
  title={FKD-Med: Privacy-Aware, Communication-Optimized Medical Image Segmentation via Federated Learning and Model Lightweighting through Knowledge Distillation},
  author={Sun, Guanqun and Shu, Han and Shao, Feihe and Racharak, Teeradaj and Kong, Weikun and Pan, Yizhi and Dong, Jingjing and Wang, Shuang and Nguyen, Le-Minh and Xin, Junyi},
  journal={IEEE Access},
  year={2024},
  publisher={IEEE}
}

@inproceedings{peketi2023flwgan,
  title={FLWGAN: Federated Learning with Wasserstein Generative Adversarial Network for Brain Tumor Segmentation},
  author={Peketi, Divya and Chalavadi, Vishnu and Mohan, C Krishna and Chen, Yen Wei},
  booktitle={2023 International Joint Conference on Neural Networks (IJCNN)},
  pages={1--8},
  year={2023},
  organization={IEEE}
}

@inproceedings{thonglek2024hierarchical,
  title={Hierarchical Federated Learning for Predicting Water Levels: A Case Study in Thailand},
  author={Thonglek, Kundjanasith and Maipradit, Arnan},
  booktitle={2024 16th International Conference on Computer and Automation Engineering (ICCAE)},
  pages={218--222},
  year={2024},
  organization={IEEE}
}

@Article{zhu2023privacy,
  title={Privacy-preserving federated learning of remote sensing image classification with dishonest majority},
  author={Zhu, Jiang and Wu, Jun and Bashir, Ali Kashif and Pan, Qianqian and Yang, Wu},
  journal={IEEE Journal of Selected Topics in Applied Earth Observations and Remote Sensing},
  volume={16},
  pages={4685--4698},
  year={2023},
  publisher={IEEE}
}

@Article{gu2022research,
  title={Research on Carbon Foam Image Segmentation Based on Deep Learning},
  author={Gu, Peng and Zhang, Ping and Jiang, Mingfei and Qiu, Xiaofeng and Wang, Xinyan and Du, Chun and Peng, Tianyou},
  journal={Wireless Communications and Mobile Computing},
  volume={2022},
  number={1},
  pages={5965302},
  year={2022},
  publisher={Wiley Online Library}
}

@Article{yu2020segmentation,
  title={Segmentation of river scenes based on water surface reflection mechanism},
  author={Yu, Jie and Lin, Youxin and Zhu, Yanni and Xu, Wenxin and Hou, Dibo and Huang, Pingjie and Zhang, Guangxin},
  journal={Applied Sciences},
  volume={10},
  number={7},
  pages={2471},
  year={2020},
  publisher={MDPI}
}

@Article{liang2020waternet,
title={WaterNet: An adaptive matching pipeline for segmenting water with volatile appearance},
author={Liang, Yongqing and Jafari, Navid and Luo, Xing and Chen, Qin and Cao, Yanpeng and Li, Xin},
journal={Computational Visual Media},
pages={1--14},
year={2020},
publisher={Springer}
}

@inproceedings{sauer2024adversarial,
  title={Adversarial diffusion distillation},
  author={Sauer, Axel and Lorenz, Dominik and Blattmann, Andreas and Rombach, Robin},
  booktitle={European Conference on Computer Vision},
  pages={87--103},
  year={2024},
  organization={Springer}
}

@inproceedings{chen2018encoder,
  title={Encoder-decoder with atrous separable convolution for semantic image segmentation},
  author={Chen, Liang-Chieh and Zhu, Yukun and Papandreou, George and Schroff, Florian and Adam, Hartwig},
  booktitle={Proceedings of the European conference on computer vision (ECCV)},
  pages={801--818},
  year={2018}
}

@Article{zhaksylyk2025rp,
  title={RP-SAM2: Refining Point Prompts for Stable Surgical Instrument Segmentation},
  author={Zhaksylyk, Nuren and Almakky, Ibrahim and Paranjape, Jay and Vedula, S Swaroop and Sikder, Shameema and Patel, Vishal M and Yaqub, Mohammad},
  journal={arXiv preprint arXiv:2504.07117},
  year={2025}
}

@Article{beutel2020flower,
  title={Flower: A Friendly Federated Learning Research Framework},
  author={Beutel, Daniel J and Topal, Taner and Mathur, Akhil and Qiu, Xinchi and Fernandez-Marques, Javier and Gao, Yan and Sani, Lorenzo and Kwing, Hei Li and Parcollet, Titouan and Gusmão, Pedro PB de and Lane, Nicholas D},
  journal={arXiv preprint arXiv:2007.14390},
  year={2020}
}

@inproceedings{wolf-etal-2020-transformers,
    title = "Transformers: State-of-the-Art Natural Language Processing",
    author = "Thomas Wolf and Lysandre Debut and Victor Sanh and Julien Chaumond and Clement Delangue and Anthony Moi and Pierric Cistac and Tim Rault and Rémi Louf and Morgan Funtowicz and Joe Davison and Sam Shleifer and Patrick von Platen and Clara Ma and Yacine Jernite and Julien Plu and Canwen Xu and Teven Le Scao and Sylvain Gugger and Mariama Drame and Quentin Lhoest and Alexander M. Rush",
    booktitle = "Proceedings of the 2020 Conference on Empirical Methods in Natural Language Processing: System Demonstrations",
    month = oct,
    year = "2020",
    address = "Online",
    publisher = "Association for Computational Linguistics",
    url = "https://www.aclweb.org/anthology/2020.emnlp-demos.6",
    pages = "38--45"
}

@Article{info11020125,
    AUTHOR = {Buslaev, Alexander and Iglovikov, Vladimir I. and Khvedchenya, Eugene and Parinov, Alex and Druzhinin, Mikhail and Kalinin, Alexandr A.},
    TITLE = {Albumentations: Fast and Flexible Image Augmentations},
    JOURNAL = {Information},
    VOLUME = {11},
    YEAR = {2020},
    NUMBER = {2},
    Article-NUMBER = {125},
    URL = {https://www.mdpi.com/2078-2489/11/2/125},
    ISSN = {2078-2489},
    DOI = {10.3390/info11020125}
}

\end{document}